\documentclass{article}

\usepackage[nonatbib, final]{neurips_2022}
\usepackage[numbers]{natbib}

\usepackage[utf8]{inputenc} 
\usepackage[T1]{fontenc}    
\usepackage{hyperref}       
\usepackage{url}            
\usepackage{booktabs}       
\usepackage{amsfonts}       
\usepackage{nicefrac}       
\usepackage{microtype}      
\usepackage{xcolor}         

\usepackage{amsmath}
\usepackage{multirow}
\usepackage{graphicx}

\usepackage{algorithm, algorithmic}
\usepackage{amsthm,amsmath,amssymb}
\usepackage{color}
\newtheorem*{remark}{Remark}

\newtheorem{definition}{Definition}
\usepackage{subfigure}
\usepackage{bm}
\usepackage{cleveref}
\usepackage{enumitem}
\usepackage{wrapfig}
\usepackage{amsmath}
\usepackage{tablefootnote}
\usepackage{soul}

\title{Dynamic Sparse Network for Time Series Classification: Learning What to “See”}

\author{%
  Qiao Xiao$^{1,*}$, Boqian Wu$^{2,}$\thanks{Equal contribution.}  , Yu Zhang $^{3,4}$, Shiwei Liu$^{5,1}$, Mykola Pechenizkiy$^{1}$, \and \textbf{Elena Mocanu$^{2}$, Decebal Constantin Mocanu $^{1,2}$} \\
  $^1$ Eindhoven University of Technology, $^2$ University of Twente\\
  $^3$ Southern University of Science and Technology\\
  $^4$ Peng Cheng Laboratory, $^5$ University of Texas at Austin\\
  \texttt{\{q.xiao,m.pechenizkiy\}@tue.nl} \\
  \texttt{\{b.wu,e.mocanu,d.c.mocanu\}@utwente.nl} \\
  \texttt{yu.zhang.ust@gmail.com} \\
  \texttt{shiwei.liu@austin.utexas.edu} \\
}

\begin{document}

\maketitle

\begin{abstract}
The receptive field (RF), {which determines the region of time series to be “seen” and used}, is critical to improve the performance for time series classification (TSC).
However, the variation of signal scales across and within time series data, makes it challenging to decide on proper RF sizes for TSC.
In this paper, we propose a dynamic sparse network (DSN) with sparse connections for TSC, which can learn to cover various RF without cumbersome hyper-parameters tuning. The kernels in each sparse layer are sparse and can be explored under the constraint regions by dynamic sparse training, which makes it possible to reduce the resource cost. 
The experimental results show that the proposed DSN model can achieve state-of-art performance on both univariate and multivariate TSC datasets with less than 50\% computational cost compared with recent baseline methods, opening the path towards more accurate resource-aware methods for time series analyses. 
Our code is publicly available at: \url{https://github.com/QiaoXiao7282/DSN}.

\end{abstract}

\section{Introduction}

Time series classification (TSC) as an important research topic in the data mining communities \cite{ismail2019deep}, has a wide range of applications from health monitoring \cite{ismail2019automatic, forestier2018surgical}, and public security \cite{yi2018integrated}, to grid energy \cite{mocanu2018line}, and remote sensing \cite{pelletier2019temporal}. In the last decade, neural networks, especially deep neural networks, have achieved competitive or even better performance than traditional TSC approaches (e.g. DTW \cite{rakthanmanon2012searching}, BOSS \cite{schafer2015boss}, COTE \cite{bagnall2015time}) in many cases \cite{wang2017time, ismail2019deep}. 

\begin{figure}[!htb]
    \centering
    \includegraphics[width=\textwidth]{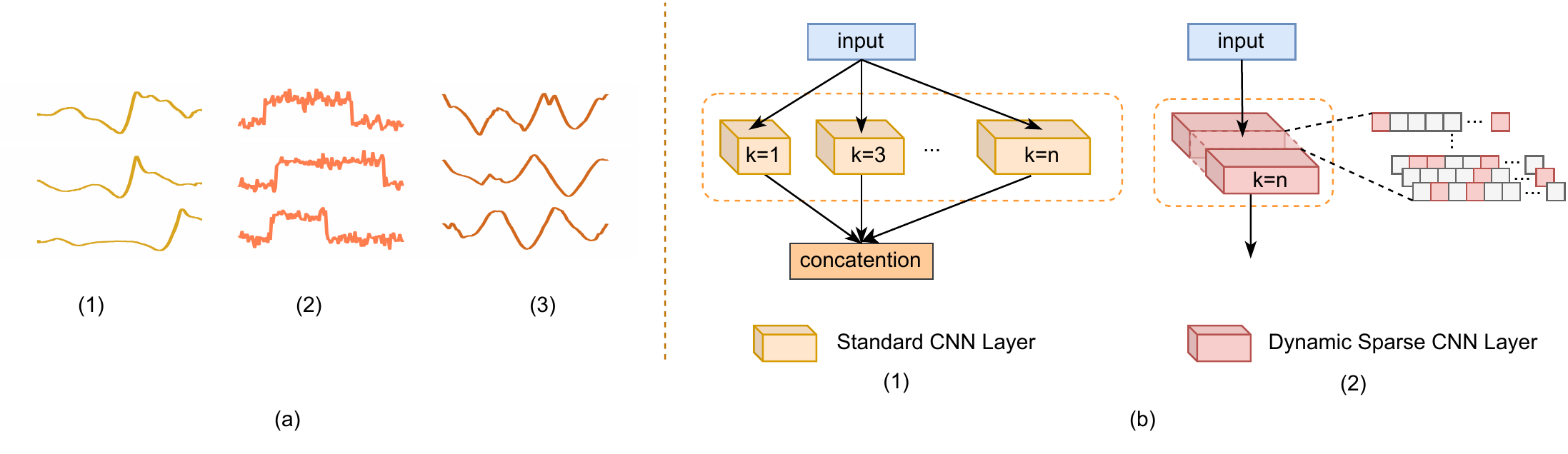}
    \vskip -0.1in
    \caption{(a) Toy example illustrates the various time scales characteristic of TS data. Data in (1), (2), and (3) are sampled from three different TS datasets. The sequences with the same color represent the same class. We can find that several sources of variances like phase, warping and offset are common in TS data. (b) The comparison between other methods \cite{tang2020rethinking, ismail2020inceptiontime} (shown in b1) and our proposed DSN (shown in b2) for TSC. To cover various RF, DSN does not need to be configured with a series of kernel sizes $k$.}
    \label{fig:introduction}
    \vskip -0.05in
\end{figure}

However, how to discover and exploit the various scaled signals hidden in time series (TS) is still a significant challenge for TSC tasks \cite{tang2021omni, LucasSPOZGPW19, HillsLBMB14}. One main reason lies in several variances, like sampling rate and record length, which are natural during time-series collection \cite{schafer2015boss, lines2012shapelet, dau2019ucr}. Furthermore, the amplitude offset, warping and occlusion of data points are unavoidable (see Figure~\ref{fig:introduction}). So determining the optimal scales for feature extraction is difficult but significant in the TSC task.
One main solution is to cover as many receptive fields (RF) scales as possible, so it does not ignore any useful signals from the time series inputs \cite{tang2021omni, dempster2020rocket, Rocket, ismail2020inceptiontime}.
%

Grounding on the motivation that multiple kernels in combination can capture discriminative patterns despite complex warping, Fawaz et al. \cite{fawaz2019deep} tried to ensemble convolutional neural networks with a large number and variety of kernels. Tang et al. \cite{tang2020rethinking} have stacked several OS-CNN blocks, each of which consisted of a list of kernel sizes to cover all scales of RF and achieve better performance on several TSC tasks. In Multi-scale Convolutional Neural Network (MCNN) \cite{cui2016multi}, a grid search was applied to find suitable kernel sizes. Without elaborate setting and fine-grid search, Dempster et al. \cite{dempster2020rocket} transformed time series by using random convolutional kernels and then trained a time series classifier. Even though the use of kernels with various sizes can help to extract latent hierarchical features of multiple resolutions, it tends to cause a huge computational problem and overfits the datasets when a limited amount of training data is available \cite{ismail2020inceptiontime}.
To address this problem, dilated convolution \cite{yu2015multi} has been used to reduce the computational cost while keeping the receptive field size. Franceschi et al. \cite{franceschi2019unsupervised} presented an unsupervised learning model with convolutional kernels for time series feature transformation, in which the dilation factors of kernels increased exponentially layer by layer. To date, there is still an unresolved challenge with hyperparameter selections (e.g. kernel sizes and dilated factors) to capture patterns at different scales. 
This raises a foundational question: \textbf{\textit{Can we design a scalable method for TSC that achieves a trade-off between computation and performance without cumbersome hyperparameter tuning?}}


Inspired by sparse neural network models which can match the performance of dense counterparts with fewer connections \cite{mocanu2018scalable, liu2021we, liu2021sparse, yuan2021mest}, we propose a dynamic sparse network (DSN) that is composed of CNN layers with large but dynamic sparse kernel for TSC tasks, which can automatically learn sparse connections in terms of covering various RF. The proposed DSN is trained by a dynamic sparse training (DST) strategy, which makes it possible to reduce the computational cost (e.g. floating-point operations (FLOPs)) for training and inference as well. Different from the traditional dynamic sparse training methods, which explore the connections in a layer-wise manner \cite{mocanu2018scalable, LiuMPP21, EvciGMCE20} and can hardly cover smaller RF, we propose a more fine-grained sparse training strategy for TSC. Specifically, the CNN kernels in each layer are split into several groups, each of which is sparse and can be explored under a constraint region during sparse training. In this way, each layer can extract features in a more scalable and diverse way, and then by cascading, the model can cover more various scales of RF. 
To sum up, our main contributions are:
\begin{itemize}

\item We propose a novel dynamic sparse CNN layer with various effective neighbour RF, which can learn to ``see'' selectively and adaptively for more accurate TSC. 
\item By stacking the above dynamic sparse CNN layers, we introduce a novel DSN model natively for TSC, which is trained by DST to reduce computational and memory cost.
\item To bridge the gap between resource awareness and accuracy, we propose a more fine-grained training strategy for DSN, which can more easily cover diverse effective RF for TSC.
\item We perform extensive empirical evaluation on both univariate and multivariate TSC tasks, showing that DSN achieves superior performance in terms of accuracy and resource cost.
\end{itemize}

\section{Related Work}
\subsection{Time Series Classification}

In the last decade, the success of deep learning encourages researchers to explore and extend its application on TSC tasks \cite{langkvist2014review, ismail2019deep, wang2017time, huang2020residual, lee2021learnable}. For univariate TSC, deep learning-based models attempt to directly transform raw time series data into low-dimensional feature representation via convolutional or recurrent networks \cite{ismail2020inceptiontime, rahimian2019xceptiontime, ouyang2021convolutional}. 
For multivariate TSC, LSTM and attention layers are often stacked with CNN layers to extract complementary features from TS data \cite{karim2019multivariate, zhang2020tapnet}. 

Recently, many works have attempted to extract features in a more scalable manner as TS data is unavoidably composed of signals on various scales \cite{schafer2015boss, dau2019ucr, tang2021omni}. One main solution is to configure suitable kernels of various sizes such that they increase the probability of capturing the proper scales \cite{tang2021omni, tang2020rethinking,  ismail2020inceptiontime}.  Rocket-based methods \cite{dempster2020rocket, Rocket, DempsterSW21} aim to use random kernels with several sizes and dilation factors to cover diverse RF for TSC. 
Different from these works, our proposed dynamic sparse CNN layer can adaptively learn the various RF with deformable dilation factors and trade-off both computation and performance without cumbersome hyperparameter tuning.

\subsection{Sparse Training}

Recently, the lottery ticket hypothesis has shown that it is  possible to train a sparse sub-network to match the performance of its dense counterpart with less computational cost \cite{frankle2018lottery}. Rather than iteratively pruning from a dense network \cite{GirishMGCDS21, ZhangCCW21, frankle2018lottery}, recent works try to find an initial mask with one-shot pruning based on gradient information during training \cite{LeeAT19, wang2019picking}. After pruning, the topology of the neural network will be fixed during training. However, this kind of models can hardly match the accuracy achieved by their dense counterparts \cite{mocanu2016topological, wang2019picking, tanaka2020pruning}.

Introduced as a new training paradigm before the lottery ticket hypothesis, DST starts from a sparse network and allows the sparse connectivity to be evolved dynamically with a fixed number of parameters throughout training \cite{mocanu2018scalable, EvciGMCE20, liu2021we, yuan2021mest, dettmers2019sparse}. 
Nowadays, DST attracts increasing attentions from other research fields such as reinforcement learning \cite{sokar2021dynamic} and continual learning \cite{Powerpropagation}, due to its potential of outperforming dense neural networks training \cite{mocanu2018scalable, yuan2021mest, liu2021we}. 
Different from the traditional DST methods, our proposed DSN is trained with a fine-grained sparse training strategy rather than a traditional layer-wise manner to capture more diverse RF for TSC.


\subsection{Adaptive Receptive Field}
The RF, which can be adaptively changed during the training, has proven to be effective in many domains \cite{DCN, Resolution-Learning, Context-Deformable}. The adaptive RF can usually be captured by learning the optimal kernel sizes or the masks of kernels during the course of training \cite{Spatially-Adaptive-Filter, Optimizing-Filter, Context-Deformable}. However, neither the kernels nor the masks are sparse, which may cause the huge computational problem when larger RF is needed. 
Different from these methods, our dynamic sparse CNN layers in DSN can be trained with DST, which can learn to capture variable RF with deformable dilation factors. What's more, the kernels are always sparse during training and inference, which makes it possible to reduce the computational cost.

\section{The Proposed Model}

\subsection{Problem Definition}

\begin{definition}
(\textbf{Time Series Classification (TSC)}). Given a TS instance $\mathbf{X} = \left\{\mathbf{X}_{1}, \ldots,\mathbf{X}_{n}\right\} \in \mathbb{R}^{n \times m}$, where $m$ denotes the number of variates and $n$ denotes the number of time steps, TSC aims to accurately predict the class label $y \in\{1, \ldots, c\}$ from $c$ classes. When $m$ equals 1, TSC is univariate and otherwise it is multivariate.
\end{definition}

\begin{definition}
(\textbf{Time Series (TS) Training Set}). A training set $\mathcal{D} = \{\left(\mathbf{X}^{(1)}, y^{(1)}\right), \ldots,$ $\left(\mathbf{X}^{(N)}, y^{(N)}\right)\}$ consists of $N$ time series instances, where $\mathbf{X}^{(i)}\in \mathbb{R}^{n \times m}$ could either be a univariate or multivariate time series instance with its corresponding label $y^{(i)} \in\{1, \ldots, c\}$.
\end{definition}

Note that, in our case, all instances have the same number of time steps in a TS dataset. Without the loss of generality, given a training set, we aim to train a CNN classifier 
with adaptive and various RF with a low resource cost (e.g. memory and computation) for the TSC task.

\subsection{Dynamic Sparse CNN Layer with Adaptive Receptive Field}

A straightforward strategy to cover various RF is to apply multi-sized kernels in each CNN layer, but it has several limitations. Firstly, TS instances from different TS datasets do not have the same length and cycles \footnote{The term `cycle' means that the data exhibit rising and falling that are not of a fixed frequency.} in most cases, making it difficult to set a fixed kernel configuration for all the datasets even with prior knowledge. Secondly, obtaining a large receptive field commonly requires a large kernel or stacking more layers, which introduces more parameters and thus increases storage and computational costs.

To tackle those challenges, the proposed dynamic sparse CNN layer possesses large but sparse kernels, which are learnable to capture adaptive RF. Specifically, given the input feature map $x^l \in \mathbb{R}^{c_{l-1} \times h \times w}$ in the $l$th layer ($h$ equals 1 for univariate TSC and $c_{l-1}$ is the number of input channels) , and kernel weights $\Theta^l \in \mathbb{R}^{c_{l-1} \times c_l \times 1 \times k}$ ($k$ is the kernel size and $c_{l}$ is the number of output channels), the convolution with stride $1$ and padding in our proposed dynamic sparse CNN layer is formulated as
\vskip -0.1in
\begin{equation}
    \label{sparse convultion operation}
    \mathbf{O}_{j}= \sum_{0<i \leq c_{l-1}, i\in \mathbb{Z}} \left(\mathbf{I}^l(\Theta^l)_{i, j} \odot \Theta^l_{i, j}\right) \cdot x^l_{i},
\end{equation}
\vskip -0.1in

where $\mathbf{O}_{j} \in \mathbb{R}^{h \times w}$ denotes the output feature representation at the $j$-th output channel, $\mathbb{Z}$ denotes the set of integers, $\mathbf{I}^l(\cdot):\mathbb{R}^{c_{l-1} \times c_{l} \times 1 \times k} \rightarrow \{0,1\}^{c_{l-1} \times c_{l} \times 1 \times k}$ is an indicator function, $\mathbf{I}^l(\Theta^l)_{i,j}$ indicates activated weights for $\Theta^l_{i, j} \in \mathbb{R}^{1 \times k}$ which is the kernel in the $(i, j)$-th channel, $\odot$ denotes the element-wise product, and $\cdot $ denotes the convolution operator. The indicator function $\mathbf{I}^l(\cdot)$, which is learned during the training of the proposed DSN, satisfies that $\|\mathbf{I}^l(\cdot)\|_0 \leq (1-S)N_l$ where $0 \leq S<1$ is the sparsity ratio, $\|.\|_0$ denotes the $L_0$ norm and $N_l = c_{l-1} \times c_{l} \times 1 \times k$. When $S>0$, the kernel is sparse, and we can use a large $k$ in the dynamic sparse CNN layer to obtain a large RF with reduced computational cost. 

\begin{wrapfigure}{r}{0.5\textwidth}
\vskip -0.4in
\begin{minipage}{0.5\textwidth}
\begin{center}
    \includegraphics[width=\textwidth]{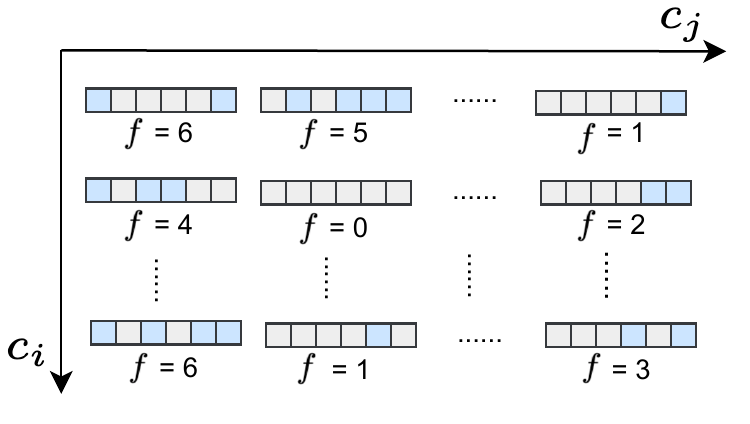}
\end{center}
\vskip -0.2in
\caption{Illustrations for the {eNRF} size. The weights with blue color are activated.} 
\label{fig:RF}
\end{minipage}
\end{wrapfigure}

\begin{remark}
(\textbf{Effective {Neighbour} Receptive Field (e{NRF}) size in the dynamic sparse CNN layer}). The receptive field is defined as the region in the input that the feature of a CNN model is looking at. {We defined the Neighbour Receptive Field (NRF) as the region considered by each feature in the successive layer. Specifically, the NRF size is equivalent to the kernel size in the standard CNN layer (Considering the case that dilation equals to 1)}. Differently, the {NRF} size of the proposed dynamic sparse CNN layer is smaller than the kernel size, when the first or last weight in a kernel is not activated, e.g., $ \exists i \in \{1,\ldots,c_{l-1}\}$ and $j \in \{1,\ldots,c_{l}\}$, such that $I^l(\Theta^l)_{i,j,1,1} = 0 $ or $I^l(\Theta^l)_{i,j,1,k} = 0$. 
As illustrated in Figure~\ref{fig:RF}, we defined the {eNRF} size $f^l_{i,j}$ of the kernel $\Theta^l_{i,j}$ as the distance between the first activated weight and last activated weight in the $l$-th CNN layer as
\begin{equation}
f^l_{i,j} := \left\{\begin{array}{ll} \max(\textit{Ind}^l_{i,j})-\min(\textit{Ind}^l_{i,j}) + 1, & \mathrm{if}\ \textit{Ind}^l_{i,j} \neq \varnothing \\
   0, & \mathrm{otherwise}
   \end{array}\right.,
\end{equation}
where $\textit{Ind}^l_{i,j}$ denotes the set of indices corresponding to non-zero weights in the kernel $\Theta^l_{i,j}$. Obviously we have $ 0 \leq f^l_{i,j} \leq k$.
\end{remark}

The {eNRF} size set of $l$-th dynamic sparse CNN layer is denoted by $\mathbb{F}^{(l)}$, which satisfies that $0 \leq \min(\mathbb{F}^{(l)})$ and $\max(\mathbb{F}^{(l)}) \leq k$. Take the case in the Figure~\ref{fig:RF} as an simple example for $l$-th layer, then $\mathbb{F}^{(l)} = \{0, 1, 2, 3, 4, 5, 6\}$. Each dynamic sparse CNN layer is likely to cover various {eNRF} sizes ranging from $1$ to $k$.
When global information is expected, $\mathbf{I}^l(\cdot)$ could activate weights more dispersed to get a larger {eNRF} and exploit the input features selectively. To capture the local context, a smaller {eNRF} is expected, thus the activated weights tend to be concentrated. By taking $k=5$ as an example, $\mathbf{I}^l(\Theta^l)_{i,j}$ may be [1, 0, 1, 0, 1] for global context, while it may be [0, 0, 1, 1, 0] for local context. In
this way, the {eNRF} can be adaptively adjusted.

\subsection{Architecture in DSN}

\begin{figure}[htb]
\begin{minipage}{\textwidth}
\centering
\includegraphics[width=0.9\textwidth]{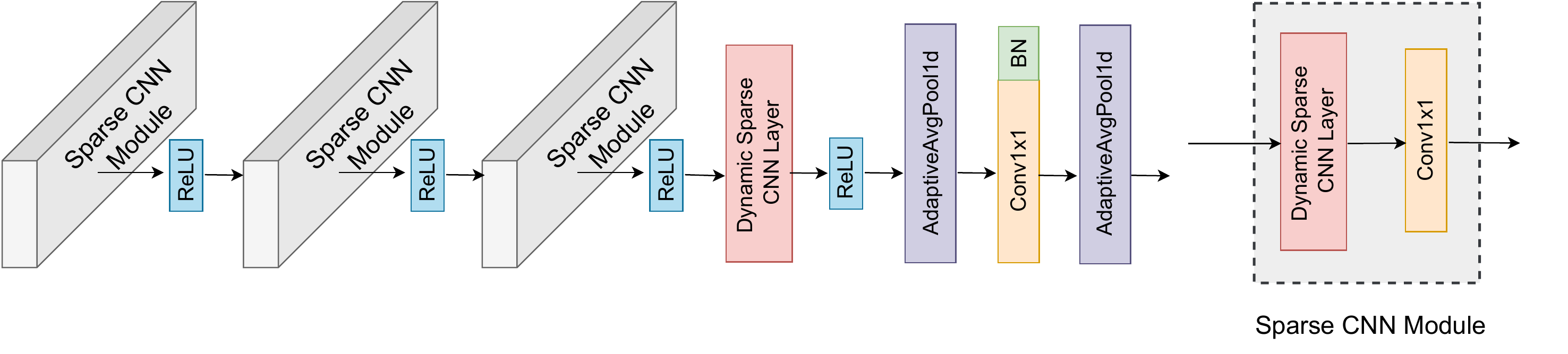}
\caption{The proposed DSN model (left), which includes several sparse CNN modules (right) followed by a dynamic sparse CNN layer, two adaptive average pooling layers, and one $1 \times 1$ convolution layer.}
\label{fig:overall network}
\end{minipage}
\end{figure}

The proposed DSN model consists of three sparse CNN modules, each of which composes of a dynamic sparse CNN layer and a $1\times1$ CNN layer. Following the stacked sparse CNN modules, there is an additional dynamic sparse CNN layer, two adaptive average pooling layers, and one $1\times1$ convolution layer acting as the classifier in the DSN model. The overall architecture is shown in Figure~\ref{fig:overall network}.

The set of {eNRF} sizes $\mathbb{S}^{(l)}$ in the $l$th sparse CNN module is equal to that in the dynamic sparse CNN layer, since the {eNRF} size of $1 \times 1$ convolution is constantly equal to $1$. $\mathbb{S}^{(l)}$
satisfies that $0 \leq \min(\mathbb{S}^{(l)})$ and $\max(\mathbb{S}^{(l)}) \leq k$.
Then, the set of {eNRF} sizes for the three successively stacked sparse CNN modules can be described by $\mathbb{RF}$ as

\begin{equation}
\mathbb{RF} =\left\{\max(0, {s}^{(1)}+{s}^{(2)}+{s}^{(3)}-2) \mid s^{(i)} \in \mathbb{S}^{(i)}, i \in\{1,2,3\}\right\}.\label{equ_RF_def}
\end{equation}

According to Eq.~\eqref{equ_RF_def}, we can see that stacking multiple sparse CNN modules can increase the size of $\mathbb{RF}$ linearly, where {$l$th} dynamic sparse CNN layer increases by the size of $\mathbb{S}^{(l)}$. For simplicity, the kernel size in each dynamic sparse CNN layer is consistently set to $k$ in our study. Then, $\mathbb{RF}$ satisfies that $\max(\mathbb{RF}) \leq 3k-2$ and $0 \leq \min(\mathbb{RF})$. Thus, a large $k$ in sparse CNN modules can increase the range of the {eNRF} sizes for stacked sparse CNN modules. 

\begin{figure}[!htb]
    \centering
    \includegraphics[width=0.7\textwidth]{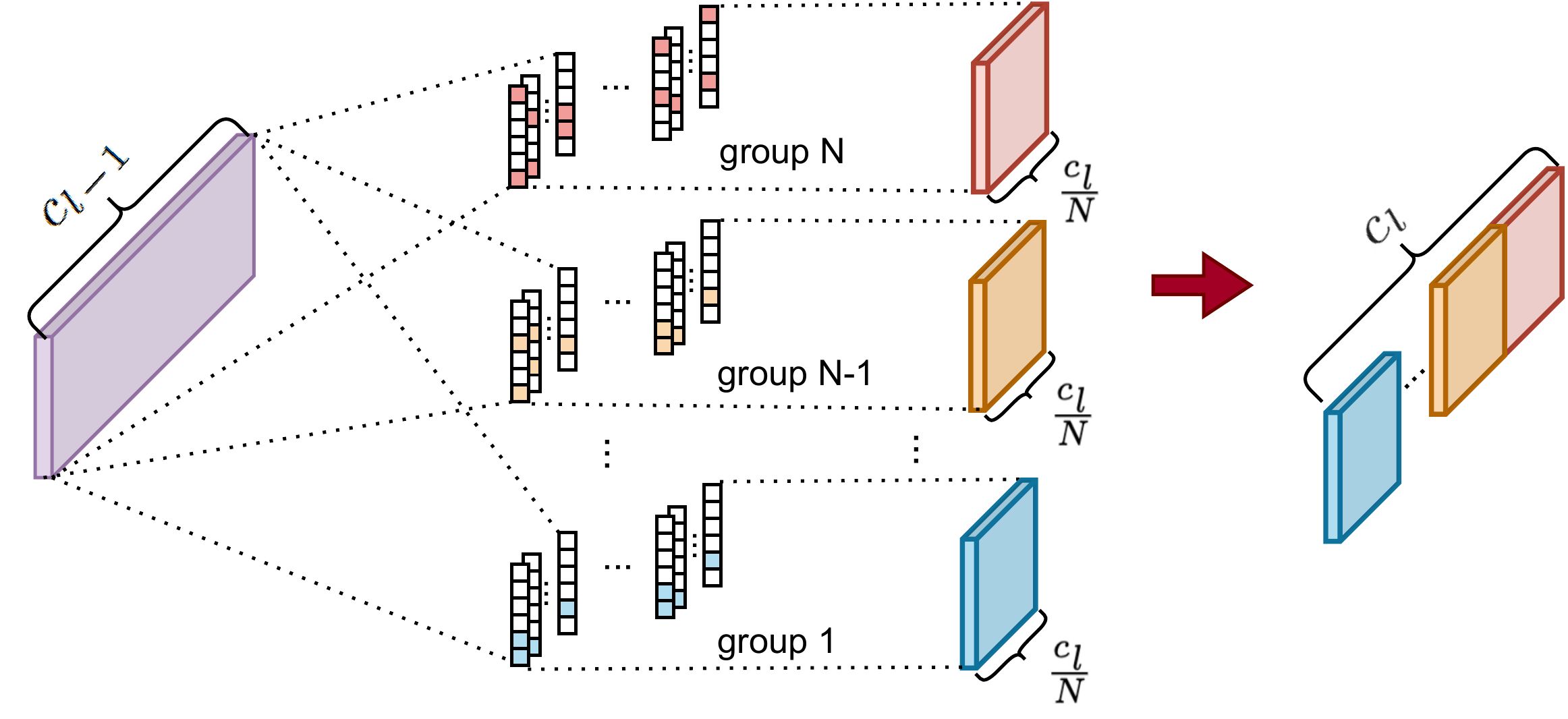}
    \caption{An illustration of the dynamic sparse CNN layer. In a dynamic sparse CNN layer, the kernels are split into $N$ groups, each of which is with sparse connections learned by dynamic sparse training across specific constraint regions. The connections with color indicate non-zero values.}
    \label{fig:sparse CNN module}
    \vskip -0.05in
\end{figure}

\subsection{Dynamic Sparse Training for DSN Model}
\label{section:3.4}

In this section, we present the training strategy to discover the weights required to be activated in order to ensure well-performing RF. That is, we need to study how to update the indicator function $\mathbf{I}^l(\cdot)$ during the training of the proposed DSN.

\renewcommand{\algorithmicrequire}{\textbf{Input:}}
\begin{wrapfigure}{r}{0.55\textwidth}
\vskip -0.15in
\begin{minipage}{0.55\textwidth}
\begin{algorithm}[H]\small
\label{alg1}
\caption{Dynamic Sparse Training for DSN Model}
\label{alg:DSN}
\begin{algorithmic}[1]
\REQUIRE Dataset $\mathcal{D}$, Network $f_{\Theta}$, Sparsity ratio $S$\\
Exploration regions in layer $l$: $\{\mathcal{R}^l_1, ..., \mathcal{R}^l_N\}$\\
Exploration schedule: $T$, $\Delta T$, $\alpha$, $f_{\text {decay }}$ \\  
\STATE $\theta^l_i \leftarrow$ initialize activated weights in $\Theta^l_i $ using $S$ and $\mathcal{R}^l_i$
\FOR{$t=1$ {\bfseries to} $T$}
\STATE Sample a batch  $B_{t} \sim \mathcal{D}$ \\
\STATE $L_{t}=\sum_{i \sim B_{t}} L\left(\left(f_{\theta}\left(x_{i}\right), y_{i}\right)\right.$
\IF{$(t \bmod \Delta T)==0$}
\FOR{each layer $l=1$ {\bfseries to} $L$}
\FOR{each group $i=1$ {\bfseries to} $N$}
\STATE $u= n(\mathcal{R}^l_i)f_{\text {decay}}\left(t ; \alpha, T\right)(1-S)$
\STATE $\mathbb{I}_{\text {prune }}=\operatorname{ArgTopK}\left(-\left|\theta^{l}_i\right|, u\right)$
\STATE $\mathbb{I}_{\text {grow }}=\operatorname{RandomK} \left(\mathcal{R}^l_i \backslash \theta^{l}_i, u\right)$
\STATE $\mathbf{I}^l(.) \leftarrow$ Update  $\mathbf{I}^l(.)$ using $\mathbb{I}_{\text {prune }}$ and $\mathbb{I}_{\text {grow }}$
\STATE ${\theta}^l_i \leftarrow$ Update activated weights $\mathbf{I}^l(\Theta^l)_i \odot \Theta^l_i$
\ENDFOR
\ENDFOR
\ELSE 
\STATE $\theta=\theta-\alpha \nabla_{\theta} L_{t}$
\ENDIF
\ENDFOR
\end{algorithmic}
\end{algorithm}
\end{minipage}
\end{wrapfigure}

We sparsely train the proposed DSN model from scratch to keep kernel sparse following the main idea of DST methods \cite{mocanu2018scalable}. By design, during the training phase, the total number of activated weights must not exceed $N_l(1-S)$. However, we observe that the tiny {eNRF} is hard to be captured by DST methods with a layer-wise exploration manner, namely activated weights are discovered layer by layer, especially when the sparsity ratio $S$ is undersized (more analysis in Section \ref{section:4.5}). Motivated by this observation, the kernels in each dynamic sparse CNN layer are divided into different groups, whose corresponding exploration regions are of different sizes as shown in Figure~\ref{fig:sparse CNN module}. Contrary to the DST methods, in DSN, the exploration of activated weights is separately performed in different kernel groups, which is a more fine-grained strategy. Specifically, the kernel weights  $\Theta^l \in \mathbb{R}^{c_{l-1} \times c_{l} \times 1 \times k}$ in the $l$th layer are split into $N$ groups along the output channel, that is, ${\Theta^l_1,\ldots, \Theta^l_N} \in \mathbb{R}^{c_{l-1} \times \frac{c_{l}}{N} \times 1 \times k}$ with corresponding exploration regions $\mathcal{R}^l_1,\ldots, \mathcal{R}^l_N$. 
The exploration region $\mathcal{R}^l_i$ in the $i$th group of the $l$th layer is defined as the first $\frac{i \times k}{N}$ positions of each kernel in this group. By taking $k=6$ and $N=3$ for example, activated weights in the first group are only within the first two positions of each kernel in $\Theta^l_1$, while they are within the whole kernel for the last group (as shown in Figure~\ref{fig:sparse CNN module}). In this way, the various {eNRF} can be covered, and exploration space can be reduced (more details can be found in Appendix \ref{appendix:A1}) to improve exploration efficiency.

Given the weight exploration regions $\mathcal{R}^l_1,\ldots, \mathcal{R}^l_N$ and the sparsity ratio $S$, we train the proposed DSN model as shown in Algorithm \ref{alg:DSN}. The activated weights are explored within the exploration regions and updated every $\Delta_t$ iterations. The fraction of updated weight decays over time according the function $f_{\text {decay}}\left(t ; \alpha, T\right)$, which follows the cosine annealing \cite{Dettmerscosin} as
\vskip -0.15in
\begin{equation}
\label{eq:cosine annealing}
f_{decay}\left(t ; \alpha, T\right)=\frac{\alpha}{2}\left(1+\cos \left(\frac{t \pi}{T}\right)\right),
\end{equation}
\vskip -0.1in
where $\alpha$ is the initial fraction of activated weights updated, $t$ is the current training iteration, and $T$ is the number of training iterations. Thus, during the $t$th iteration, the number of updated activated weights in the $i$th group of the $l$th layer is $n(\mathcal{R}^l_i)f_{\text {decay}}\left(t ; \alpha, T\right)(1-S)$, where $n(\mathcal{R}^l_i)$ is the number of weights can be explored in region ${R}^l_i$.
During the update of activated weights, we firstly prune the activated weights determined by $\operatorname{ArgTopK}\left(-\left|\theta^{l}_i\right|, u\right)$ and then randomly grow new activated weights by $\operatorname{RandomK} \left(\mathcal{R}^l_i \backslash \theta^{l}_i, u\right)$,  where $\operatorname{ArgTopK}(\mathbf{v}, u)$ gives the indices of top $u$ elements with larger values in a vector $\mathbf{v}$, $\operatorname{RandomK}(\mathbf{v}, u)$ outputs the indices of random $u$ elements in a vector $\mathbf{v}$, and $\mathcal{R}^l_i \backslash \theta^{l}_i$ denotes the weights within $\mathcal{R}^l_i$ but except $\theta^{l}_i$. 

The exploration of activated weights is straightforward. Firstly, pruning weights with small magnitudes is intuitive, because the contribution of weights with smaller magnitudes is insignificant or even negligible. Considering the pruning recoverability, 
we randomly regrow new weights with the same number as the pruned weights to achieve better activated weight exploration. In this way, the exploration of weights is dynamic and of plasticity, {compared with the methods of pruning weights before and after training (more explanations about pruning and dynamic sparse training can be found in \cite{10.5555/3546258.3546499, 10.5555/3463952.3463960, Shiwei})}. 

\section{Experiments}
In this section, we evaluate the resource cost and accuracy of the proposed DSN model with recent baseline methods on both univariate and multivariate TSC data. 

\subsection{Datasets and Baselines}

The details of each kind of dataset are as follows.
\begin{itemize}
\item \textbf{Univariate TS dataset from UCR 85 archive \cite{UCRArchive}}: This archive consists of 85 univariate TS datasets, which are collected from various domains (e.g. health monitoring and remote sensing), have distinguishable characteristics, and various levels of complexity. The number of instances in the training set varies from 16 to 8926, while their time step resolutions range from 24 to 2709.  

\item \textbf{Three multivariate TS datasets from UCI \cite{Dua:2019}}: The EEG2 dataset contains 1200 instances of 2 categories and 64 variates. The Human Activity Recognition (HAR) dataset consists of 10,299 instances with 6 categories, and the number of variates is 9. The Daily Sport dataset contains 9,120 instances of 19 categories, and its number of variates is 45. The processing of these datasets is the same as in \cite{karim2019multivariate}.
\end{itemize}

\textbf{Univariate TSC baselines:} The performance of the following three baseline methods are reported, OS-CNN \cite{tang2021omni}, InceptionTime \cite{ismail2020inceptiontime}, and ResNet\cite{wang2017time}, as they are widely used for univariate TSC. 
\textbf{Multivariate TSC baselines:} Similar with \cite{zhang2020tapnet}, the following three multivariate baselines are selected: OS-CNN, TapNet \cite{zhang2020tapnet}, MLSTM-FCN \cite{karim2019multivariate}.
To avoid an unfair comparison, we let outside the ensemble methods. However, based on our best knowledge, our performance analysis accounts for most of state-of-the-art methods.

\subsection{Experimental Settings and Implementation Details}
\label{exp_detail}
For optimization, we use the Adam optimizer with an initial learning rate of $3\times 10^{-4}$, cosine decayed to $10^{-4}$. Our model is trained for 1,000 epochs with a mini-batch size of 16. As in  \cite{ismail2020inceptiontime, karim2019multivariate}, the best model, which corresponds to the minimum training loss, is used to evaluate performance over the testing sets. Inspired by \cite{ismail2020inceptiontime}, the kernel sizes $k$ in each dynamic sparse CNN layer are set to 39. The number of kernel groups $N$ in Algorithm \ref{alg:DSN} is set to 3, which helps to cover the small, medium, and large eNRF. Each setting is repeated five times, and the average results are reported.

\subsection{Results and Analysis}
We show the performance on both univariate and multivariate TSC benchmarks in Table \ref{tab:UCRresults}-\ref{tab:multi}. For univariate TSC, we can see that our method outperforms the baseline methods in most cases with a smaller number of parameters (e.g. Params \footnote{The amount of non-zero parameters in the model as in \cite{sokar2021dynamic, KusupatiRSW0KF20, LeeAT19}.}) and less computation cost (e.g. FLOPs) on UCR 85 archive datasets. For multivariate TSC, our proposed DSN method achieves better performance on EEG2 and HAR datasets, while for Daily Sport, MLSTM-FCN was with 0.46\% more accurate. It is worth highlighting that the resource cost (i.e. Params and FLOPs) of our proposed DSN is much smaller than that of baseline methods. Due to the page limitation, the detailed results, including test accuracy and resource cost on each dataset of UCR 85 archive, can be found in Appendix \ref{appendix:C}. More results on UCR 128 \cite{UCRArchive2018} and UEA 30 archive \cite{abs-1811-00075} can be found in Appendix \ref{appendix:B7}

\begin{table}[!htb]
\vskip -0.05in
\caption{Pairwise comparison of test accuracy (\%) and mean resource cost (i.e. Params ($K$)$\downarrow$ and FLOPs ($M$)$\downarrow$) of our method and other univariate TS baseline methods on UCR 85 archive datasets. Note that all the accuracies in comparison are rounded to two decimal places.}
\centering
\resizebox{0.98\linewidth}{!}{
\begin{tabular}{c|c|ccccc}
\toprule[1pt]
Archive& Methods & Baseline wins &  \textbf{DSN} wins &  Tie & Params & FLOPs \\
\toprule[0.5pt] \bottomrule[0.5pt] 
\multirow{5}{*}{UCR 85 archive} &\multicolumn{1}{l|}{ResNet} &29 &\textbf{53} &3 &479.20 &803.38 \\
& \multicolumn{1}{l|}{InceptionTime } &\textbf{40} &\textbf{40} &5 &389.34 &325.49 \\
& \multicolumn{1}{l|}{OS-CNN } &33 &\textbf{49} &3 &262.14 &200.12 \\
& \multicolumn{1}{l|}{\textbf{DSN (ours)}} &- &- &- &\textbf{126.22} &\textbf{104.88} \\
\bottomrule[1pt]
 \end{tabular}
 }
 \label{tab:UCRresults}
\vskip -0.05in
\end{table}

\begin{table}[!htb]
\centering
\caption{Test accuracy (ACC(\%)) and resource cost (i.e. Params ($K$)$\downarrow$ and FLOPs ($M$)$\downarrow$) of different models on three multivariate TS datasets from UCI.}
\renewcommand\arraystretch{1.2}
\resizebox{\linewidth}{!}{
\begin{tabular}{l|rrr|rrr|rrr}
\toprule[1pt]
&  \multicolumn{3}{c|}{Daily Sport} & \multicolumn{3}{c|}{EEG2} & \multicolumn{3}{c}{HAR}  \\  
\cline{1-10} 
Methods & ACC& Params& FLOPs& ACC& Params& FLOPs& ACC& Params& FLOPs    \\  
\toprule[0.5pt] \bottomrule[0.5pt] 
TapNet  & 99.56& 1885.58& 346.97& 85.30& 1369.21& 780.19& 93.33& 1275.78&310.61 \\ 
MLSTM-FCN  & \textbf{99.65}& 324.77& 70.59& 91.00& 342.51& 161.53& 96.71& 284.97&63.38 \\ 
OS-CNN  & 99.30& 317.05& 78.01& 93.03& 320.34& 162.04& 96.37& 280.39&70.98 \\ 
\textbf{DSN (ours)}  & 99.19& \textbf{163.68}& \textbf{40.48}& \textbf{99.10}& \textbf{162.14}& \textbf{82.31}& \textbf{96.82}& \textbf{160.13}&\textbf{40.64} \\ 
\bottomrule[1pt]
\end{tabular}
}
\label{tab:multi}
\end{table}

\subsection{Sensitivity Analysis}
\textbf{Effect of Sparsity Ratio:}
From Table \ref{tab:density_UCI}, we can find out how the sparsity ratio of dynamic sparse CNN layer affects the final test accuracy and the resource cost in multivariate TS datasets. Here, we analyze the trade-off between test accuracy and the resource cost under various sparsity ratios (i.e. $S \in [50\%, 80\%, 90\%]$) and a dense DSN model. Note that the dense model is defined exactly as our DSN model with dense connections along entire kernels. We can see that it is not that the more parameters of the model, the better performance it will achieve. This is mainly because more parameters will lead to overfitting and the proper sparse ratio will easily cover the desired receptive field (more demonstrations can be seen in Appendix~\ref{appendix:B1}). Taking the EEG2 data set as an example, with the decrease of sparse rate, the performance on it drops sharply. We hypothesise that this data set has more expectations for features from small receptive fields (the validation can be found in Appendix~\ref{appendix:B2}).
According to the experiments, a 80\% sparsity ratio reflects a good trade-off between accuracy and resource cost, which is the default setting for the dynamic sparse CNN layer in our DSN model.


\begin{table}[!htb]
\centering
\caption{Test accuracy (ACC(\%)) and resource cost (i.e. Params ($K$)$\downarrow$ and FLOPs ($M$)$\downarrow$) of the proposed DSN model with different sparsity ratios on three multivariate TS datasets from UCI.}
\renewcommand\arraystretch{1.2}
\resizebox{\linewidth}{!}{
\begin{tabular}{l|rrr|rrr|rrr}
\toprule[1pt]
&  \multicolumn{3}{c|}{Daily Sport} & \multicolumn{3}{c|}{EEG2} & \multicolumn{3}{c}{HAR}  \\  
\cline{1-10} 
Methods & ACC& Params& FLOPs& ACC& Params& FLOPs& ACC& Params& FLOPs    \\  
\toprule[0.5pt] \bottomrule[0.5pt] 
Dense  & 99.19& 1059.65& 262.10& 89.60& 1058.11& 536.17& 96.06&1056.10 &267.57 \\ 
$S=50\%$  & \textbf{99.28}& 370.44& 91.63& 96.23& 368.90& 187.05& 96.62& 366.89&93.01 \\ 
$S=80\%$  & 99.19& 163.68& 40.48& \textbf{99.10}& 162.14& 82.31& \textbf{96.82}& 160.13&40.64 \\ 
$S=90\%$  & 99.25& \textbf{94.76}& \textbf{23.44}& 98.90& \textbf{93.22}& \textbf{46.64}& 96.75& \textbf{91.21}&\textbf{23.18} \\ 
\bottomrule[1pt]
\end{tabular}
}
\label{tab:density_UCI}
\end{table}

\begin{figure}[!htb]
\vskip -0.1in
    \centering
    \includegraphics[width=0.9\textwidth]{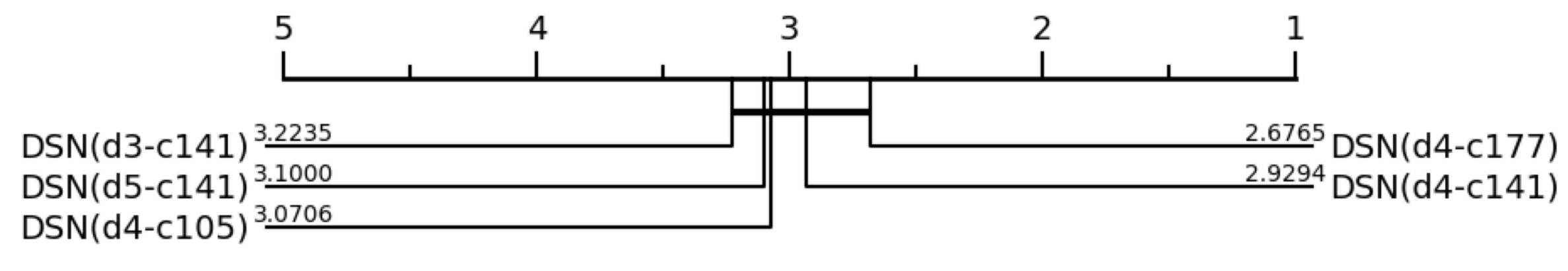}
    \vskip -0.1in
    \caption{The critical difference diagram shows the average rank of the proposed DSN model with different architectures in UCR 85 archive. DSN(d$i$-c$j$) represents the DSN model with $i$ dynamic sparse CNN layers and $j$ output channels in each layer. The smaller average rank corresponds to the better performance of the model.}
    \label{fig:sensitivity_analyse}
\end{figure}

\textbf{Effect of Architecture:}
To show how the architecture of DSN affects the final performance, we adopt the critical difference diagram for a detailed comparison, which is usually used for evaluation in TSC \cite{tang2021omni, ismail2020inceptiontime}.
From Figure~\ref{fig:sensitivity_analyse}, we can find that the DSN model with 4 dynamic sparse CNN layers and 177 output channels in each layer performs consistently better than the other structures. The proposed model with more output channels seems to be crucial for preserving the capacity of the network, while requires more parameters and FLOPs compared to other counterparts. This highlights an interesting trade-off between accuracy and computational efficiency. According to the experiments, the DSN model with 4 dynamic sparse CNN layers and 141 output channels in each layer is the trade-off between the performance and resource cost, which is the default setting for our DSN model.

\begin{figure}[!htb]
    \vskip -0.1in
    \centering
    \includegraphics[width=\textwidth]{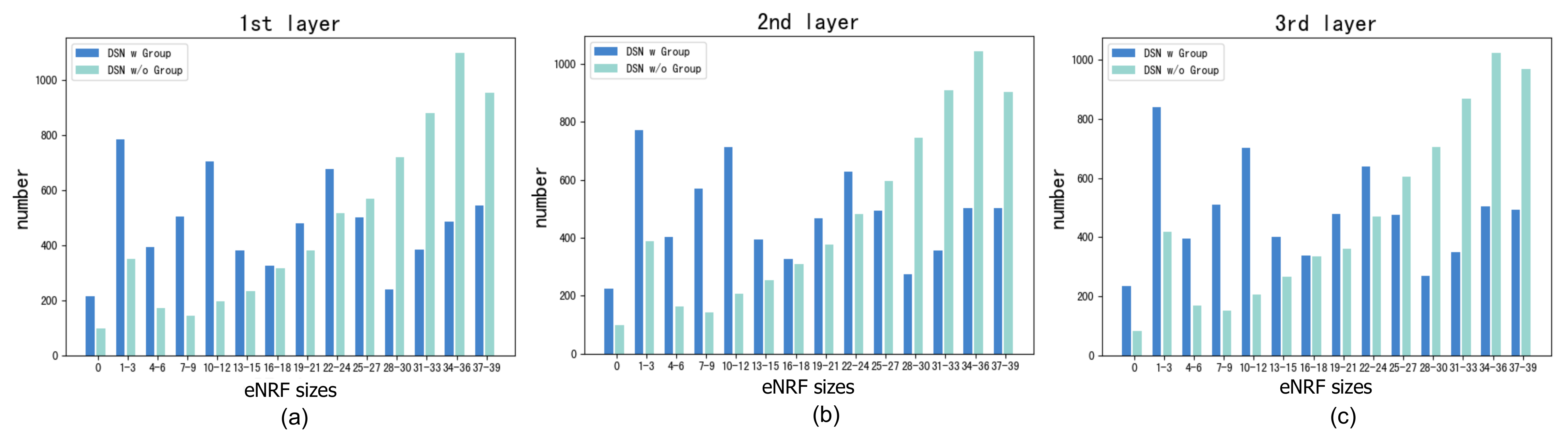}
    \vskip -0.1in
    \caption{The distribution of {eNRF} sizes in different sparse CNN layers between DSN models training with and without grouping on HAR dataset. The number on y-axis corresponds to the frequency of {eNRF} sizes.}
    \label{fig:eRF_sizes}
    \vskip -0.1in
\end{figure}

\textbf{Effect of Kernel Group:} As shown in Figure~\ref{fig:eRF_sizes}, training DSN without kernel group hampers the capture of small eNRF, which can not guarantee satisfactory performance on the datasets such as EEG2 expecting more local information. With the decrease in sparsity ratio $S$, the phenomenon that large RF occupies the majority becomes more severe (more demonstrations see in Appendix \ref{appendix:B1}), which increases the performance gap between training with group and without it, as shown in Table \ref{tab:ablation_study}. Grouping the kernels can achieve various RF coverage and adorable accuracy for all datasets.  

\subsection{Ablation Study}
\label{section:4.5}

\begin{wraptable}{r}{0.6\textwidth}
\vskip -0.25in
\begin{minipage}{0.6\textwidth}
\caption{Average accuracies (\%) of the proposed DSN model affected by kernel grouping.}
\resizebox{\linewidth}{!}{
\begin{tabular}{c|c|ccc}
\toprule[1pt]
Sparsity ratio& Methods & Daily Sport &  EEG2 &  HAR  \\
\toprule[0.5pt] 
\midrule[0.5pt]
\multirow{2}{*}{$S=60\%$} &\multicolumn{1}{l|}{DSN w Group} &\textbf{99.27} &\textbf{97.87} &96.53 \\
& \multicolumn{1}{l|}{DSN w/o Group} &99.23 &96.77 &\textbf{96.87}  \\
\midrule[0.5pt]
\multirow{2}{*}{$S=70\%$} &\multicolumn{1}{l|}{DSN w Group} &99.20 & \textbf{98.70}& \textbf{97.14} \\
& \multicolumn{1}{l|}{DSN w/o Group} &\textbf{99.25} &97.27 &96.52  \\
\midrule[0.5pt]
\multirow{2}{*}{$S=80\%$} &\multicolumn{1}{l|}{DSN w Group} &99.19 & \textbf{99.10}& \textbf{96.82} \\
& \multicolumn{1}{l|}{DSN w/o Group} &\textbf{99.28} &98.77 &96.79  \\
\bottomrule[1pt]
 \end{tabular}
 }
\label{tab:ablation_study}
\end{minipage}
\vskip -0.1in
\end{wraptable}%

\begin{figure}[H]
\vskip -0.1in
\begin{minipage}{\textwidth}
\begin{center}
    \includegraphics[width=\textwidth]{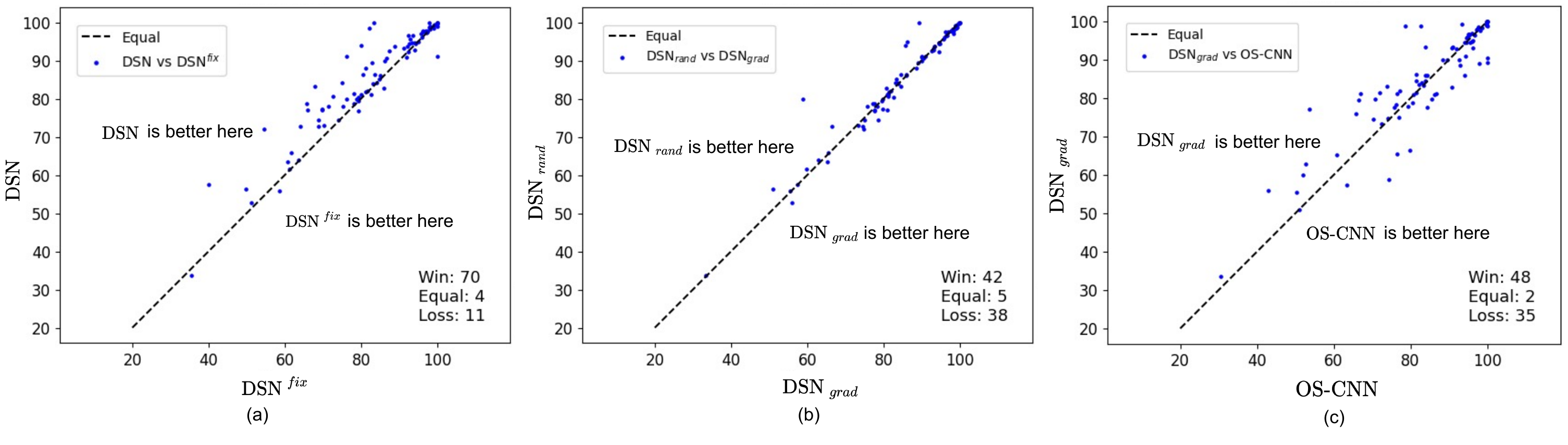}
\end{center}
\vskip -0.2in
\caption{Accuracy plotting showing how the proposed DSN model is affected by the dynamic sparse training in UCR 85 archive.}
\label{fig:ablation_study}
\end{minipage}
\vskip -0.1in
\end{figure}

\textbf{Effect of Dynamic Sparse Training:}
Furthermore, we study the effect of the dynamic sparse training in our DSN method. We conduct ablation studies to compare DSN with its static variant (i.e. DSN$^{fix}$), which fixes its topology during training after activated weights initialization. In other words, the indicator function $\mathbf{I}^l(\cdot)$ in Eq.~\eqref{sparse convultion operation} will not update during the training process. 
According to the results shown in Figure~\ref{fig:ablation_study} (a), we can find that the DSN model mostly outperforms the DSN$^{fix}$ on UCR 85 archive, indicating that what is learned by dynamic sparse training is the suitable activated weights together with their values.


\subsection{Case Study on other DST Method}

\begin{wrapfigure}{r}{0.6\textwidth}
\vskip -0.18in
\begin{minipage}{0.6\textwidth}
\begin{center}
    \includegraphics[width=\textwidth]{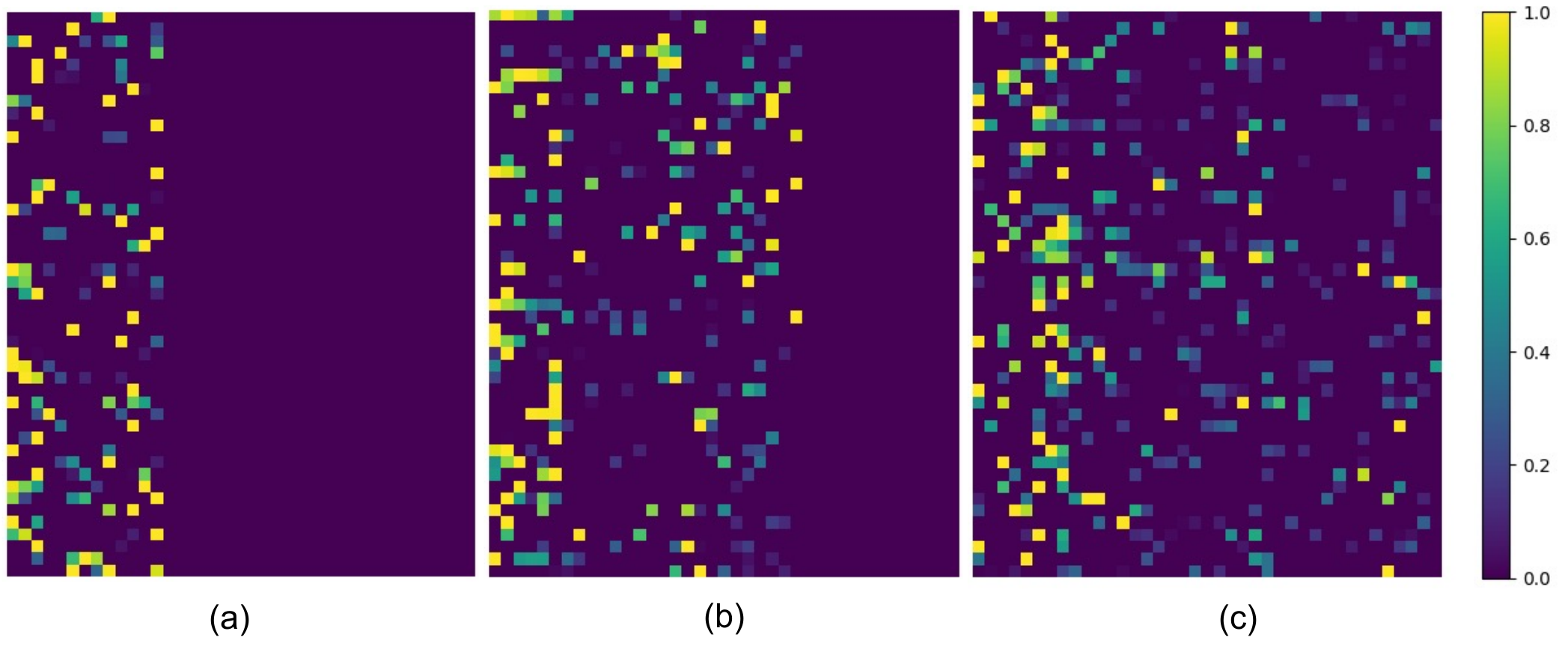}
\end{center}
\vskip -0.15in
\caption{The activated weights in different kernel groups.}
\label{fig:weights}
\end{minipage}
\vskip -0.1in
\end{wrapfigure}

Likewise, the effectiveness of the proposed DSN method is also analyzed in the presence of other DST methods, such as RigL \cite{EvciGMCE20}. Rather than growing weights randomly (as in the line 9 of Algorithm~\ref{alg:DSN}), RigL grows them with the highest magnitude gradients. 
From Figure~\ref{fig:ablation_study} (c), we can see that growing weights according to gradients (i.e. DSN$_{grad}$)
can outperforms the OS-CNN method in most cases on UCR 85 archive. 
Furthermore, as shown in Figure~\ref{fig:ablation_study} (b), we can find that DSN$_{grad}$ has a similar performance as growing weights randomly (i.e. DSN$_{rand}$). 
Note that, DSN$_{rand}$ is the main setting of the proposed method in this paper with the reason that we don't need to compute the full gradients at any moment in time.
In summary, the proposed DSN with fine-grained sparse strategy can be easily combined with the existing DST techniques, showing the potential for further improvements by merging it with other advanced DST methods in the future.

\subsection{Visualization of Activated Weights}

Figure \ref{fig:weights} shows the normalized activated weights of the DSN model after sparse training on HAR dataset. (a), (b) and (c) correspond to the activated weights of the three kernel groups in the first dynamic sparse CNN layer, and each row of the figure represents one kernel. We can see that the weights in each kernel are sparse, and can be activated within the constraint region defined in Section~\ref{section:3.4}.

\section{Conclusions}
\label{conclusion}

This paper proposes dynamic sparse network (DSN) to cover diverse effective neighbour receptive field (eNRF) for time series classification (TSC) without the cumbersome hyper-parameters tuning. The proposed DSN can achieve about $2 \times$ reduction in memory and computational cost, while achieving superior performance in terms of accuracy on both univariate and multivariate TSC datasets. Furthermore, DSN explores the activated weights in a more fine-grained strategy, which can be easily merged with existing DST techniques, showing a potential for further improvements when combined with advanced technology. DSN provides a feasible solution to bridge the gap between resource awareness and various eNRF coverage for TSC, hoping to inspire other researchers in different fields. It is worthwhile extending DSN to other research domains (e.g. time series forecasting and computer vision). Our DSN model has not yet been executed on real edge devices, and the performance in terms of accuracy and memory footprint on edge devices is not investigated. In the future, we hope to extend our work to other fields and decentralise training on edge devices.

\section*{Acknowledgements}
This work has been partly funded by the NWO Perspectief project MEGAMIND. {We thank everyone who worked on making the datasets available and their contributions to the TSC community.}

{\small
\bibliography{neurips_2022}
\bibliographystyle{plain}
}

\clearpage
\section*{Checklist}


\begin{enumerate}

\item For all authors...
\begin{enumerate}
  \item Do the main claims made in the abstract and introduction accurately reflect the paper's contributions and scope? \answerYes{}
  \item Did you describe the limitations of your work?
    \answerYes{} See Section~\ref{conclusion}
  \item Did you discuss any potential negative societal impacts of your work?
    \answerYes{} See Section~\ref{conclusion}
  \item Have you read the ethics review guidelines and ensured that your paper conforms to them?
    \answerYes{}
\end{enumerate}

\item If you are including theoretical results...
\begin{enumerate}
  \item Did you state the full set of assumptions of all theoretical results?
    \answerNA{}
        \item Did you include complete proofs of all theoretical results?
    \answerNA{}
\end{enumerate}

\item If you ran experiments...
\begin{enumerate}
  \item Did you include the code, data, and instructions needed to reproduce the main experimental results (either in the supplemental material or as a URL)?
    \answerYes{}
  \item Did you specify all the training details (e.g., data splits, hyperparameters, how they were chosen)? 
    \answerYes{} See Section~\ref{exp_detail}
        \item Did you report error bars (e.g., with respect to the random seed after running experiments multiple times)?
    \answerYes{}
        \item Did you include the total amount of compute and the type of resources used (e.g., type of GPUs, internal cluster, or cloud provider)?
    \answerYes{} See Appendix~\ref{appendix:D}
\end{enumerate}

\item If you are using existing assets (e.g., code, data, models) or curating/releasing new assets...
\begin{enumerate}
  \item If your work uses existing assets, did you cite the creators?
    \answerYes{}
  \item Did you mention the license of the assets?
    \answerNA{}
  \item Did you include any new assets either in the supplemental material or as a URL?
    \answerYes{}
  \item Did you discuss whether and how consent was obtained from people whose data you're using/curating?
    \answerNA{}
  \item Did you discuss whether the data you are using/curating contains personally identifiable information or offensive content?
    \answerNA{}
\end{enumerate}

\item If you used crowdsourcing or conducted research with human subjects...
\begin{enumerate}
  \item Did you include the full text of instructions given to participants and screenshots, if applicable?
    \answerNA{}
  \item Did you describe any potential participant risks, with links to Institutional Review Board (IRB) approvals, if applicable?
    \answerNA{}
  \item Did you include the estimated hourly wage paid to participants and the total amount spent on participant compensation?
    \answerNA{}
\end{enumerate}

\end{enumerate}


\clearpage
\appendix

\section{Preliminaries} 
\label{appendix:A}


\subsection{Preliminaries on Comparison of Exploration Space}
\label{appendix:A1}
The size of exploration space for the layer-wise manner in SET algorithm is:
\begin{equation}
\left(\begin{array}{l}
N_l \\
N_l(1-S)
\end{array}\right) = \frac{N_l!}{N_l(1-S)! N_lS!}.
\end{equation}

If we divide the kernels in one layer into $N$ groups and explore $\frac{N_l(1-S)}{N}$ activated weights in each group, then the size of the exploration space becomes:

\begin{equation}
{\left(\begin{array}{l}
\frac{N_l}{N} \\
\frac{N_l(1-S)}{N}
\end{array}\right)}^{N} = {(\frac{\frac{N_l}{N}!}{\frac{N_l(1-S)}{N}! \frac{N_lS}{N}!})}^N \leq \left(\begin{array}{l}
N_l \\
N_l(1-S)
\end{array}\right).
\end{equation}

When the corresponding exploration region of each group is defined as in Section \ref{section:3.4}, the exploration space can be further reduced to:

\begin{equation}
\left(\begin{array}{l}
\frac{N_l}{N^2} \\
\frac{N_l(1-S)}{N}
\end{array}\right)\left(\begin{array}{l}
\frac{2N_l}{N^2} \\
\frac{N_l(1-S)}{N}
\end{array}\right) \cdots\left(\begin{array}{l}
\frac{N_l}{N} \\
\frac{N_l(1-S)}{N}
\end{array}\right) < {\left(\begin{array}{l}
\frac{N_l}{N} \\
\frac{N_l(1-S)}{N}
\end{array}\right)}^N.
\end{equation}

\subsection{Preliminaries on Calculation of Parameters}
\label{appendix:A2}
The summation of the number of non-zero weights (activated weights in sparse CNN layer + weights in other layers) in the network is used to estimate the size of the network as follows:
\begin{equation}
Params =\sum_{l \notin \mathbb{L}}  \| \theta^l \|_0 +\sum_{l \in \mathbb{L}}\sum_{i=1}^{N} \|\mathbf{I}^l(\Theta^l)_i \odot \Theta^l_i\|_0 = \|\theta\|_0,
\end{equation} 
where $\mathbb{L}$ contains a set of sparse CNN layer indexes, and $\theta$ is the parameter of our DSN which includes an amount of zero values. $\|\mathbf{I}^l(\Theta^l)_i \odot \Theta^l_i\|_0$ is controlled by the sparsity ratio $S$ and the corresponding exploration regions $\mathbf{R}^l_i$.

\section{More Experiments}
\label{appendix:B}

\subsection{The Statistics of eNRF Sizes with Different Sparsity Ratios}
\label{appendix:B1}

\begin{figure}[!htb]
    \centering
    \includegraphics[width=\textwidth]{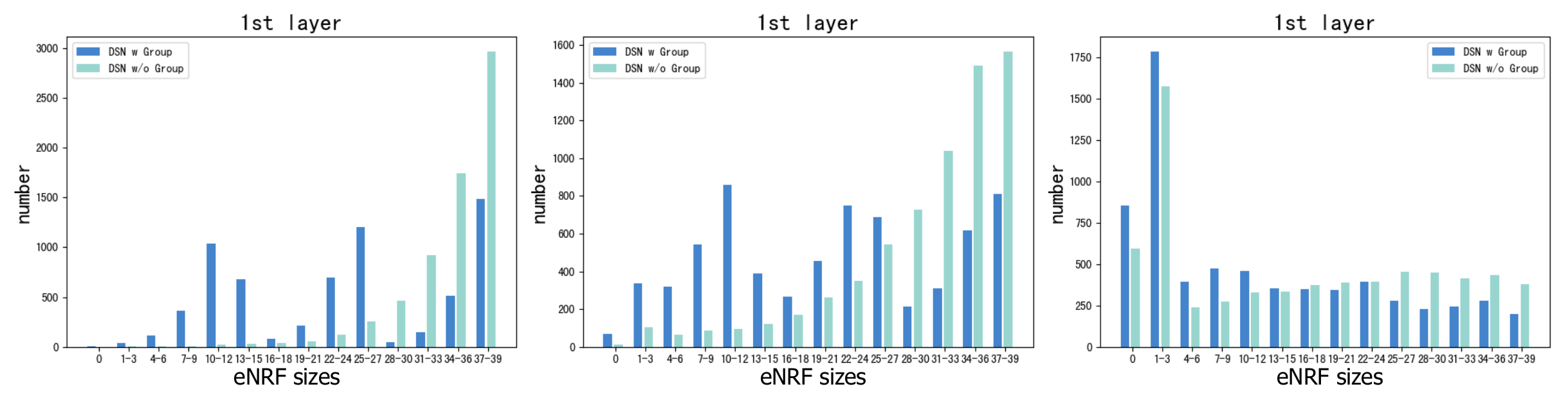}
    \caption{The statistics of eNRF sizes for the HAR dataset. The eNRF sizes shown in a sub-figure are from the first dynamic sparse CNN layer with a specific sparsity ratio (e.g.  $S=50\%$ (left), $S=70\%$ (middle), and $S=90\%$ (right)).}
    \label{fig:HAR_eRF_sizes}
\end{figure}

\begin{figure}[!htb]
    \centering
    \includegraphics[width=\textwidth]{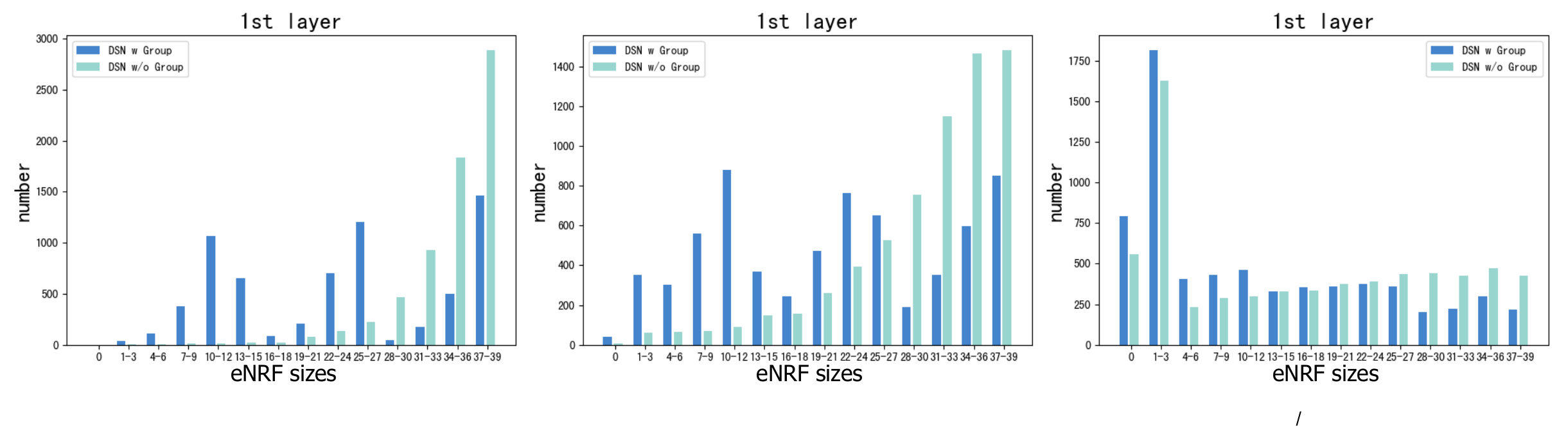}
    \caption{The statistics of eNRF sizes for the Daily Sport dataset.The eNRF sizes shown in a sub-figure are from the first dynamic sparse CNN layer with a specific sparsity ratio (e.g.  $S=50\%$ (left), $S=70\%$ (middle), and $S=90\%$ (right)).} 
    \label{fig:daily_sport_eRF_sizes}
\end{figure}

\begin{figure}[!htb]
    \centering
    \includegraphics[width=\textwidth]{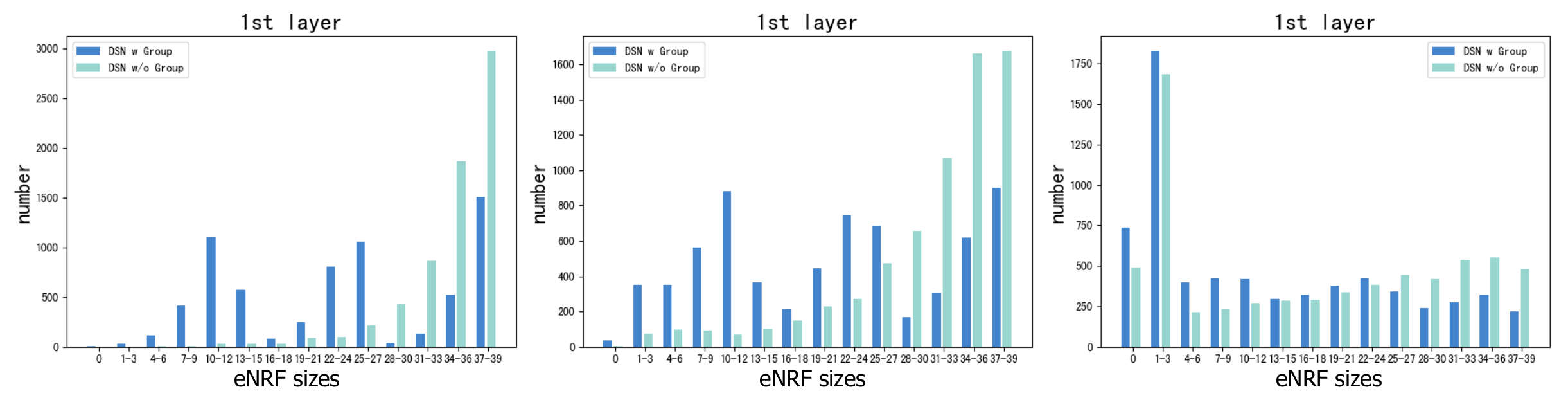}
    \caption{The statistics of eNRF sizes for the EEG2 dataset.The eNRF sizes shown in a sub-figure are from the first dynamic sparse CNN layer with a specific sparsity ratio (e.g.  $S=50\%$ (left), $S=70\%$ (middle), and $S=90\%$ (right)).} 
    \label{fig:eeg2_eRF_sizes}
\end{figure}

The distributions of eNRF sizes in the first dynamic sparse CNN layer for HAR, Daily Sport and EEG2 datasets are shown in Figure~ \ref{fig:HAR_eRF_sizes}, Figure~\ref{fig:daily_sport_eRF_sizes} and Figure~\ref{fig:eeg2_eRF_sizes}, respectively. We can see that the DSN model with kernel grouping can cover more diverse eNRF sizes than it does without kernel grouping under different sparsity ratios $S$. With the decrease of $S$, the DSN model without kernel grouping tends to capture the large eNRF, while ignoring the small eNRF, which harms the performance of the DSN model. Under different sparsity ratios, the DSN model with kernel grouping can guarantee a satisfactory performance for different datasets by covering various eNRF.  


\subsection{Better Kernel Sizes for EEG2 Dataset}
\label{appendix:B2}

\begin{figure}[!htb]
    \centering
    \includegraphics[width=0.6\textwidth]{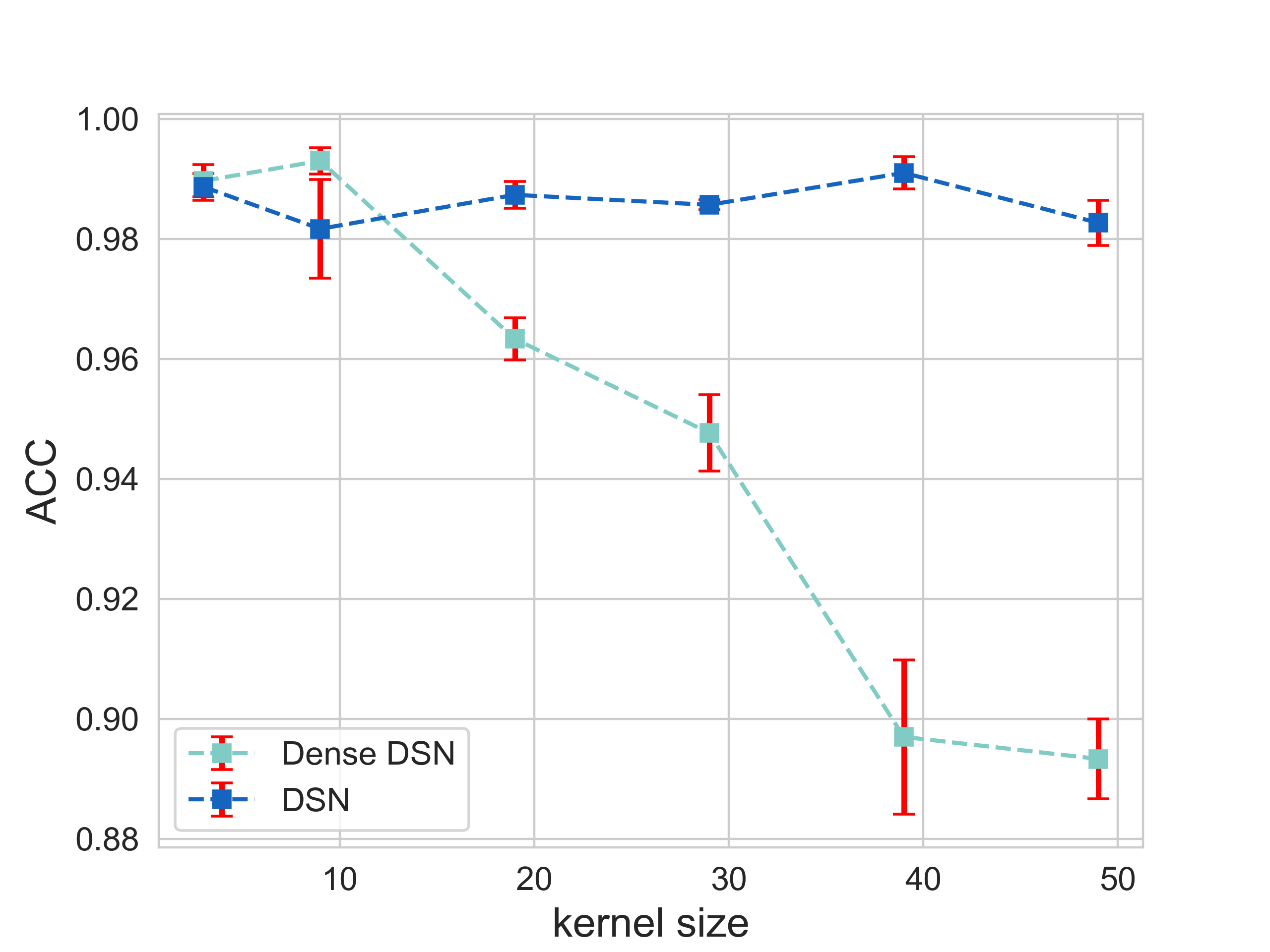}
    \caption{The test accuracies of EGG2 dataset in dense/sparse DSN with different kernel sizes $k$.}
    \label{fig:kernel_size_EGG2}
\end{figure}

We study the effect of kernel size $k$ on the performance of spares/dense DSN for the EEG2 dataset to verify our hypothesis that this dataset expects the local information. The dense DSN model is defined exactly as our DSN model with dense connections along entire kernels. In this way, the RF size in each layer equals to kernel size. As shown in Figure \ref{fig:kernel_size_EGG2}, the dense DSN with a kernel size of $9$ achieves the highest test accuracy.
However, with the increase of kernel size (RF size), the test accuracy drops dramatically. It is obvious that the local context plays a significant role for more accurate classification for EEG2 dataset. In contrast, our proposed sparse DSN model (sparsity ratio $S=80\%$) with a large kernel size can still achieve satisfactory performance, with the main reason that it can learn the adaptive eNRF for the dataset.


\subsection{Can Various eNRF Outperform the Optimal NRF?}

\begin{figure}[!htb]
    \centering
    \includegraphics[width=\textwidth]{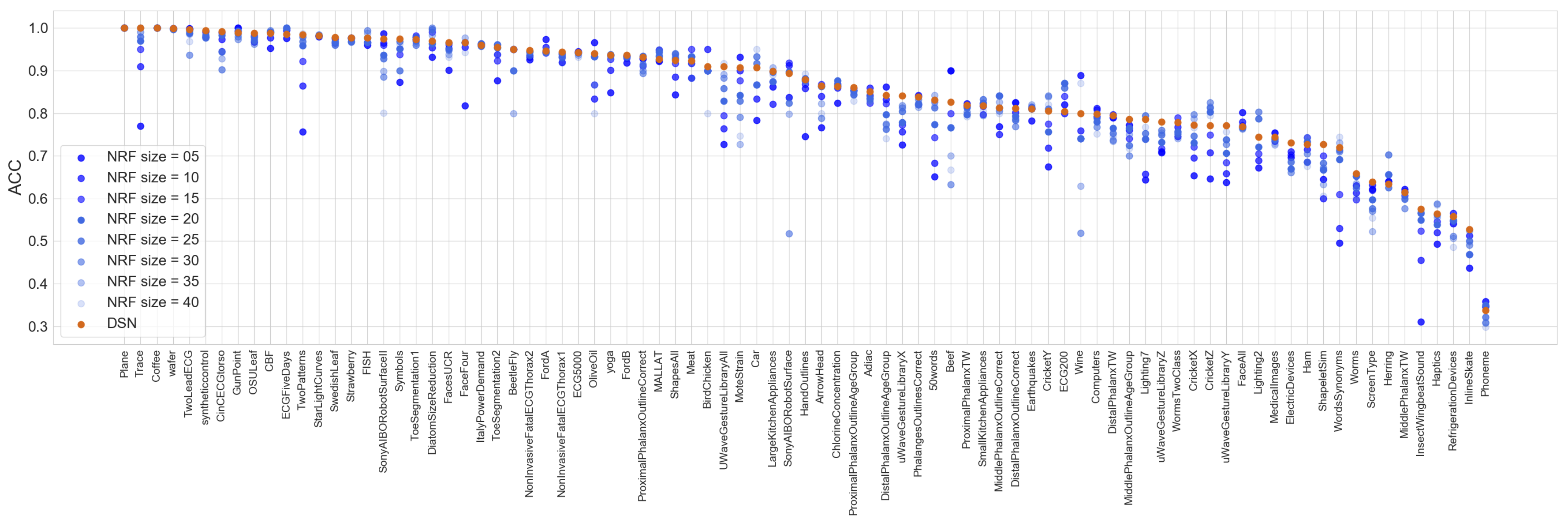}
    \caption{Classification accuracies on UCR 85 archive for the comparison of various eNRF and optimal NRF.}
    \label{fig:rf_optimal}
\end{figure}

To demonstrate the advantage of various eNRF coverage, we compare the performances in terms of test accuracy between DSN models with various eNRF coverage and optimal NRF. The dense DSN models with different kernel sizes (from 5 to 40 with step 5) in each layer are used to search for optimal NRF in different datasets on the UCR 85 archive. As shown in Figure \ref{fig:rf_optimal}, by capturing various eNRF, our DSN model can achieve similar performance to the optimal NRF model without the cumbersome hyper parameters.

\subsection{The Effect of Kernel Size for DSN}
\label{appendix:B4}
\begin{figure}[!htb]
    \centering
    \includegraphics[width=\textwidth]{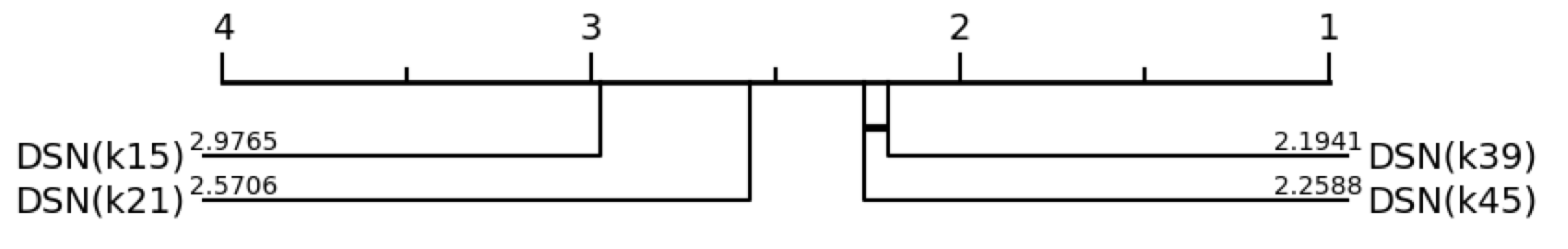}
    \caption{The critical difference diagram shows the average rank of the proposed DSN model with different kernel sizes in UCR 85 archive. DSN(k$i$) illustrates that the kernel size in each dynamic sparse CNN layer equals to $i$.}
    \label{fig:kernel_size_UCR85}
\end{figure}

From Figure~\ref{fig:kernel_size_UCR85}, we can find out how the kernel size $k$ in the proposed DSN model affects the final test accuracy for UCR 85 archive. We can see that the performance of DSN model with small kernel sizes (e.g. 15 and~21) is worse than that of it with large kernel sizes (e.g. 39 and 45). It is because the model with larger kernel sizes can cover larger eNRF, which is useful to capture global information for some datasets. However, when the kernel size is oversized (e.g. 45), some of the small eNRFs can be lost in a predefined sparsity ratio (e.g. $80\%$ in here), which will slightly degrade the performance instead.

\subsection{The Effect of Kernel Group for DSN}
\label{appendix:B5}

\begin{figure}[!htb]
    \centering
    \includegraphics[width=\textwidth]{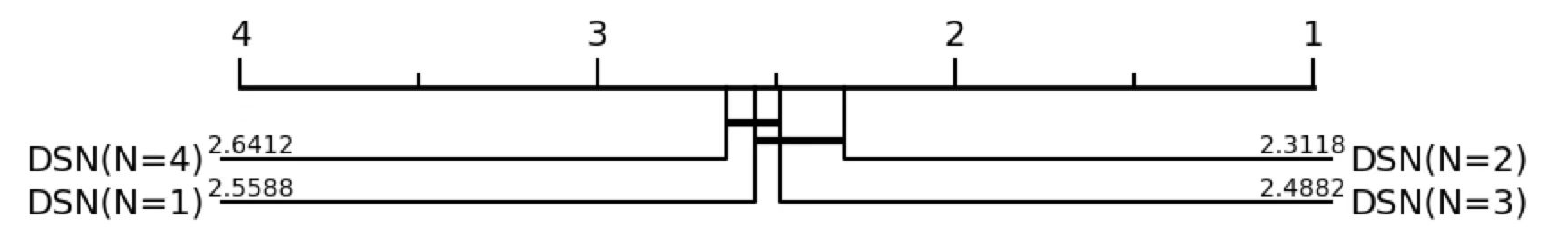}
    \caption{The critical difference diagram shows the average rank of the proposed DSN model with different kernel groups in UCR 85 archive. DSN($N=i$) illustrates that the kernels in each dynamic sparse CNN are split into $i$ groups.}
    \label{fig:group_UCR85}
\end{figure}

We study how the number of kernel groups $N$ defined in Algorithm~\ref{alg:DSN} affects the performance of the proposed DSN model. From Figure~\ref{fig:group_UCR85}, we can see that DSN model with $N=2$ outperforms other counterparts. As discussed in \ref{section:4.5} and \ref{appendix:B1}, grouping the kernels in each dynamic sparse CNN layer can help to achieve various NRF coverage especially with smaller sparsity ratio $S$. However, when the number of kernel groups (e.g. 4) is oversized, the distribution of eNRF sizes is nearly uniform, which will degrade the performance of DSN.

\begin{figure}[!htb]
    \centering
    \includegraphics[width=0.9\textwidth]{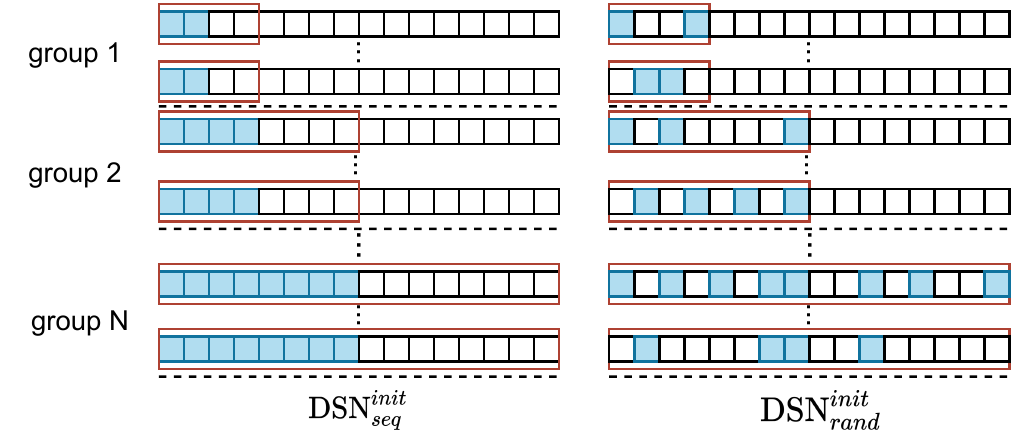}
    \caption{Visualization of different topology initialization manners, including sequential and random. The corresponding exploration regions for each group are shown in the red boxes. The blue color indicates the activated weights at initialization.}
    \label{fig: Visualizing topology initialization}
\end{figure}

\subsection{The Effect of Topology Initialization}
\begin{figure}[!htb]
    \centering
    \includegraphics[width=0.9\textwidth]{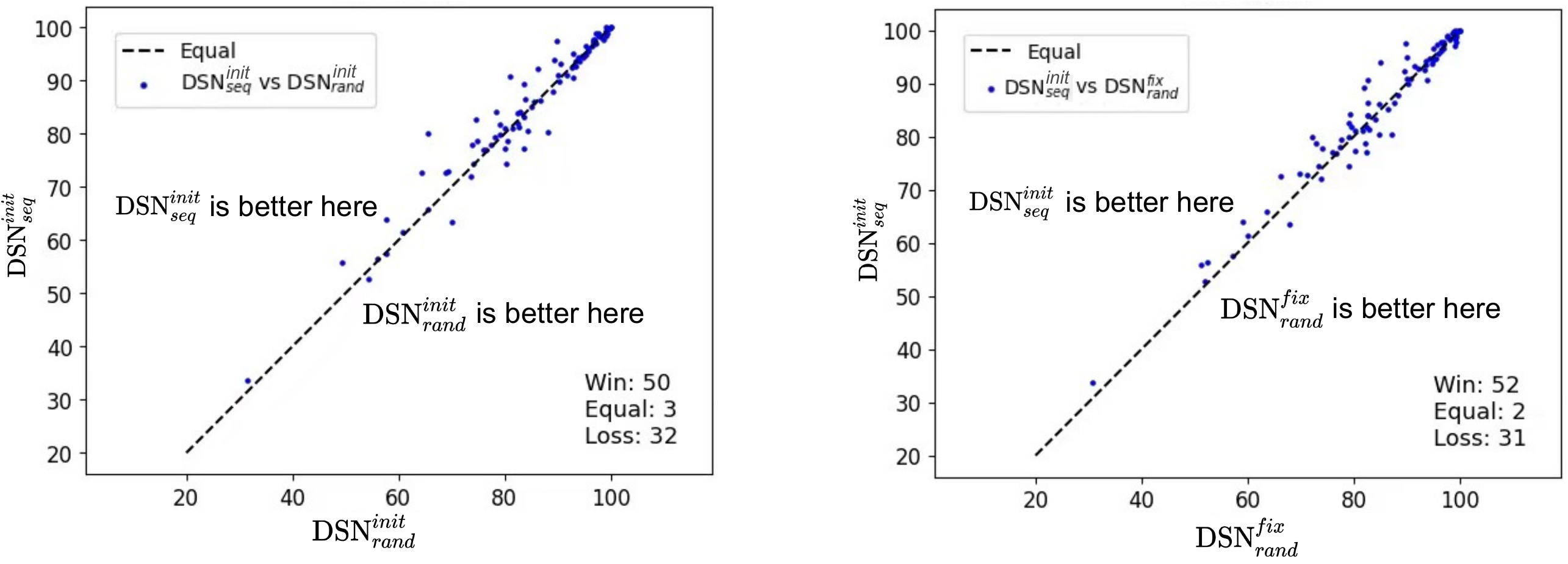}
    \caption{Accuracy plotting showing how the proposed DSN model is affected by the topology initialization on UCR 85 archive.}
    \label{fig:dst_init}
\end{figure}

\begin{figure}[!htb]
    \centering
    \includegraphics[width=\textwidth]{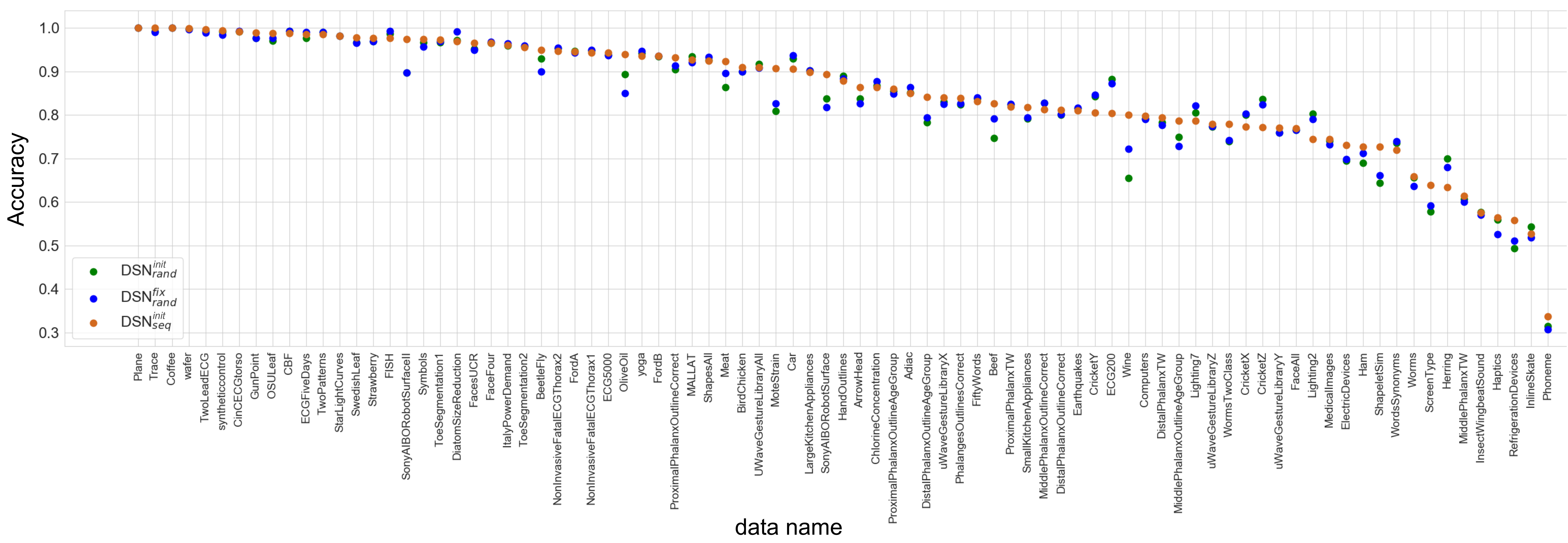}
    \caption{Classification accuracies on UCR 85 archive for the comparison of DSN$_{seq}^{init}$ and DSN$_{rand}^{init}$ with/without dynamic exploration.}
    \label{fig:acc_fix_seq}
\end{figure}

As shown in Figure~\ref{fig: Visualizing topology initialization}, the activated weights can be initialized randomly and sequentially before exploration, namely DSN$_{rand}^{init}$ and DSN$_{seq}^{init}$ respectively. After initialization, the topologies initialized by DSN$_{rand}^{init}$ and DSN$_{seq}^{init}$ are different. In detail, DSN$_{rand}^{init}$ randomly activates the weights within the exploration regions, while DSN$_{seq}^{init}$ activates the first $\frac{i \times k}{N}S$ weights of kernels in $i$th group. Figure~\ref{fig: Visualizing topology initialization} exemplifies the case where $k=16$, $N=4$ and $S=50\%$. We further investigate which type of topology initialization is more suitable for the proposed DSN model. According to the results shown in Figure~\ref{fig:dst_init} (left) and Figure~\ref{fig:acc_fix_seq}, we can see that the proposed DSN model with sequential initialization performs better than that with random initialization. Based on this, we adopt sequential initialization as the default setting for DSN model in all of the experiments.

Intuitively, the DSN method with random topology can also cover various eNRF. We further study the different performances of NRF which is from dynamic sparse training and from random topology in DSN method. 
We conduct ablation studies to compare DSN with its static variant (i.e. DSN$_{rand}^{fix}$), which fixes its topology during training after activated weights random initialization. According to the results shown in Figure~\ref{fig:dst_init} (right), we can find that the DSN model outperforms the DSN$_{rand}^{fix}$ on UCR 85 archive, indicating that what is learned by dynamic sparse training is the suitable activated weights together with their values.

\subsection{Extremely Sparse DSN}
\begin{figure}[!htb]
    \centering
    \includegraphics[width=\textwidth]{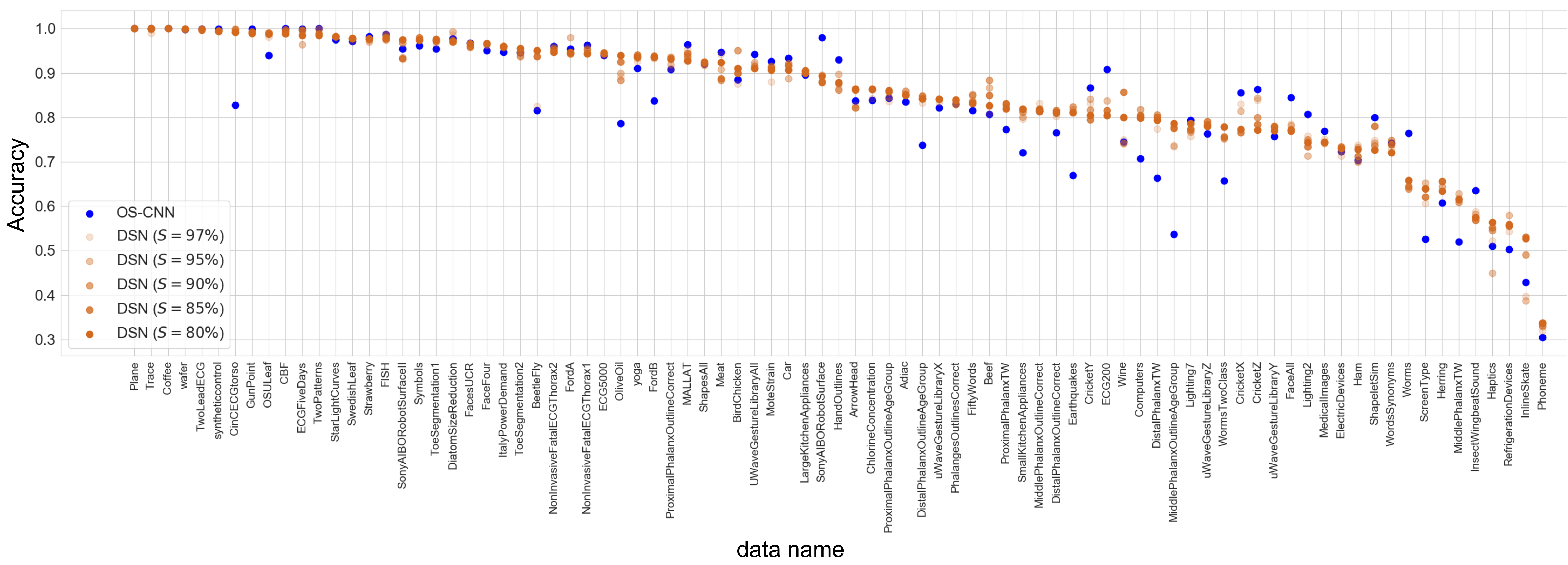}
    \caption{Classification accuracies on UCR 85 archive for DSN with different sparsity ratios.}
    \label{fig:acc_sparsity_ratios}
\end{figure}

We study how the performance of our DSN is affected under extremely sparse ratios (e.g. $S=90\%$, $S=95\%$, and $S=97\%$). As shown in Table \ref{tab:UCRresults_extremely_sparse}, we can see that our DSN with sparsity ratio $S=95\%$ still performs better than OS-CNN while achieving more than $5 \times$ resource cost reduction (e.g. Parameters and FLOPs). When sparsity ratio $S=97\%$, which means that $97\%$ weights in sparse CNN layer are set to zeros, the performance of our DSN model is only slightly worse than that of OS-CNN, but the resource cost is nearly $7 \times$ less. From Figure \ref{fig:acc_sparsity_ratios}, we can see that most blue 
dots (corresponding to accuracies of OS-CNN) are below the orange dots (corresponding to accuracies of DSN with extremely sparsity ratio), which means that OS-CNN performs worse than extremely sparse DSN in most datasets. 
What's more, DSN is far superior to OS-CNN in some datasets (e.g. CinCECGtorso, MiddlePhalanxOutlineAgeGroup, and Earthquakes). 

\label{appendix:B3}


\begin{table}[!htb]
\caption{Pairwise comparison of test accuracy (\%) and mean resource cost (i.e. Params ($K$)$\downarrow$ and FLOPs ($M$)$\downarrow$) of our method with extremely sparsity ratio and oS-CNN on UCR 85 archive datasets. Note that all the accuracies in comparison are rounded to two decimal places.}
\centering
\resizebox{\linewidth}{!}{
\begin{tabular}{c|c|ccccc}
\toprule[1pt]
Archive& Methods & OS-CNN wins & \textbf{DSN($S$)} wins &   Tie & Params & FLOPs \\
\toprule[0.5pt] \midrule[0.5pt] 
\multirow{6}{*}{UCR 85 archive} &\multicolumn{1}{l|}{OS-CNN} &- &- &- &262.14 &200.12 \\
& \multicolumn{1}{l|}{DSN($S=80\%$)} &33 &\textbf{49}  &3 &126.22 &104.88 \\
& \multicolumn{1}{l|}{DSN($S=85\%$)} &33 &\textbf{49}  &3 &100.19 &83.21 \\
& \multicolumn{1}{l|}{DSN($S=90\%$)} &36 &\textbf{47} &2 &74.16 &61.54 \\
& \multicolumn{1}{l|}{DSN($S=95\%$)} &38 &\textbf{43} &4 &48.13 &39.87 \\
& \multicolumn{1}{l|}{DSN($S=97\%$)} &\textbf{41} &40 &4 &\textbf{37.72} &\textbf{31.20} \\
\bottomrule[1pt]
 \end{tabular}
 }
 \label{tab:UCRresults_extremely_sparse}
\end{table}

\begin{figure}[!htb]
    \centering
    \includegraphics[width=\textwidth]{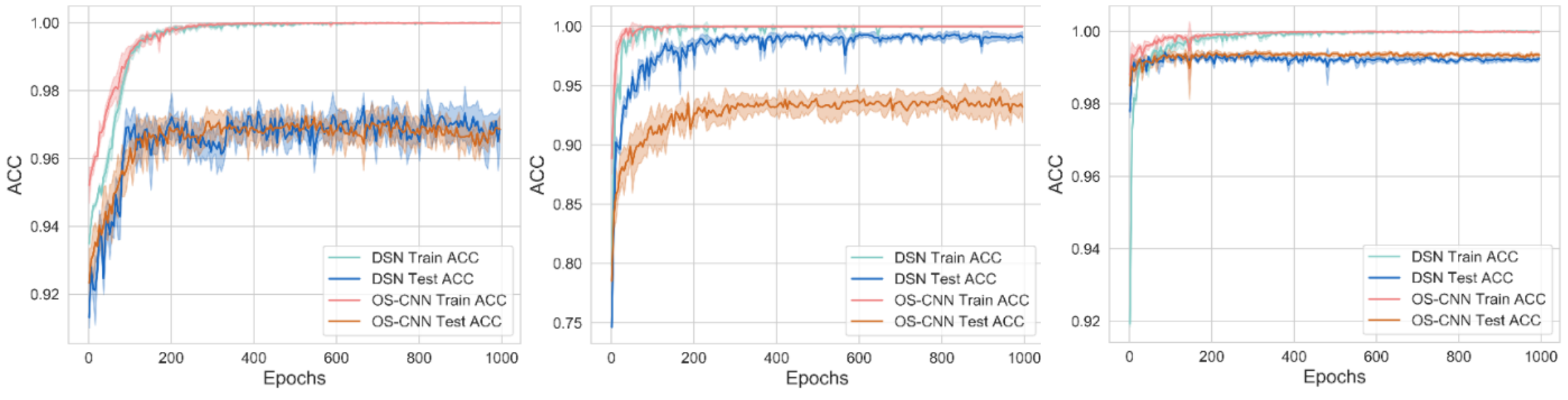}
    \caption{Learning curves of DSN models on different datasets, HAR(left), EEG2(middle), and Daily Sports(right). The standard deviation of the accuracy over 5 runs is shown in the shaded region.}
    \label{fig:learning_curve}
\end{figure}

\subsection{Learning Curve on Three Multivariate TS Dataset}
\label{appendix:B6}
Figure \ref{fig:learning_curve} shows the learning behaviors of our DSN model and OS-CNN. We can see that our DSN model has a convergence speed similar to that of OS-CNN, while the resource requirements (memory and computational cost) are much less. Furthermore, the performance of DSN is far superior to that of OS-CNN for EEG2 datasets.

{
\subsection{Additional Experiments on other TSC datasets}
\label{appendix:B7}
In order to evaluate if the good learning capabilities of DSN can be generalized over other larger scale of univariate and multivariate time-series datasets we evaluate it on the following datasets:
}

\begin{itemize}

\item {\textbf{UCR 2018 archive} \cite{UCRArchive2018}. This archive consists of 128 univariate TS datasets, which is the updated version of the UCR 85 archive. Among them, 15 datasets are with unequal lengths and one (Fungi) has a single instance per class in the training files. As shown in \cite{middlehurst2021hive}, most implementations are not set up to handle these kinds of data; therefore, we also only report our results on the remaining 112 problems in UCR 2018 archive. }

\item \textbf{University of East Anglia (UEA) 30 archive} \cite{abs-1811-00075}. It consists of 30 multivariate TS datasets with distinguishable characteristics, and various levels of complexity. The class numbers vary from 2 to 39, and their numbers of variates vary from 2 to 963.
\end{itemize}
{
For UEA 30 archive, we add the resource awareness method Rocket \cite{dempster2020rocket} and the transformer based method TST \cite{zerveas2021transformer} as our baselines.We summarize the performance on both additional univariate and multivariate TSC benchmarks in Table \ref{tab:UCR112results}-\ref{tab:UEA_results}. We can see that, in terms of test accuracy, our method can still match the state-of-the-art methods, while having much smaller computational cost (e.g. the number of parameters and FLOPs). It is worth highlighting our proposed DSN method achieves a trade-off between the resource cost and performance in both univariate and multivariate TSC benchmarks.
}

\begin{table}[!htb]
\caption{Pairwise comparison of test accuracy (\%) and mean resource cost (i.e. Params ($K$)$\downarrow$ and FLOPs ($M$)$\downarrow$) of our method and other univariate TS baseline methods on UCR 112 archive datasets. Note that all the accuracies in comparison are rounded to two decimal places.}
\centering
\resizebox{\linewidth}{!}{
\begin{tabular}{c|c|ccccc}
\toprule[1pt]
Archive& Methods & Baseline wins &  \textbf{DSN} wins &  Tie & Params & FLOPs \\
\toprule[0.5pt] \bottomrule[0.5pt] 
\multirow{5}{*}{UCR 112 archive} &\multicolumn{1}{l|}{ResNet} &43 &\textbf{64} &5 &479.28 &1049.12 \\
& \multicolumn{1}{l|}{InceptionTime } &\textbf{52} &51 &9 &389.42 &425.06 \\
& \multicolumn{1}{l|}{OS-CNN } &48 &\textbf{59} &5 &264.19 &259.30 \\
& \multicolumn{1}{l|}{\textbf{DSN (ours)}} &- &- &- &\textbf{126.30} &\textbf{136.91} \\
\bottomrule[1pt]
 \end{tabular}
 }
 \label{tab:UCR112results}
\end{table}

\begin{table}[!htb]
\caption{Pairwise comparison of test accuracy (\%) and mean resource cost (i.e. Params ($K$)$\downarrow$ and FLOPs ($M$)$\downarrow$) of our method and other multivariate TS baseline methods on UEA 30 archive datasets. Note that all the accuracies in comparison are rounded to two decimal places.}
\centering
\resizebox{\linewidth}{!}{
\begin{tabular}{c|c|ccccc}
\toprule[1pt]
Archive& Methods & Baseline wins &  \textbf{DSN} wins &  Tie & Params & FLOPs \\
\toprule[0.5pt] \bottomrule[0.5pt] 
\multirow{7}{*}{UEA 30 archive} &\multicolumn{1}{l|}{TapNet} &11 &\textbf{17} &2 &1344.67 &966.46 \\
& \multicolumn{1}{l|}{MLSTM-FCN} &7 &\textbf{23} &0 &383.41 &638.53 \\
& \multicolumn{1}{l|}{OS-CNN} &\textbf{15} &14 &1 &389.65 &547.34 \\
& \multicolumn{1}{l|}{TST} &12 &\textbf{18} &0 &343.78 &628.95 \\
& \multicolumn{1}{l|}{Rocket} &\textbf{15} &14 &1 &- &526.87\tablefootnote{For transform operation with 1000 random convolution kernels.} \\
& \multicolumn{1}{l|}{\textbf{DSN (ours)}} &- &- &- &\textbf{148.74} &\textbf{305.37} \\
\bottomrule[1pt]
 \end{tabular}
 }
 \label{tab:UEA_results}
\end{table}

\section{Detailed Results}
\label{appendix:C}

For UCR 85 archive, the reported results of ResNet and InceptionTime are from \cite{dempster2020rocket}, while that of OS-CNN are from the official repository \footnote{https://github.com/Wensi-Tang/OS-CNN.\label{fn:1}}. For three multivariate datasets from UCI, the reported results of LSTM-FCN are from the original paper, and that of TapNet and OS-CNN are from the implementation of official repository with default settings. For UEA 30 archive, the detailed results, including test accuracy and resource cost on each dataset, can be found in Table \ref{tab:UCRresults_detailed_2}.

\begin{figure}[!htb]
    \centering
    \includegraphics[width=\textwidth]{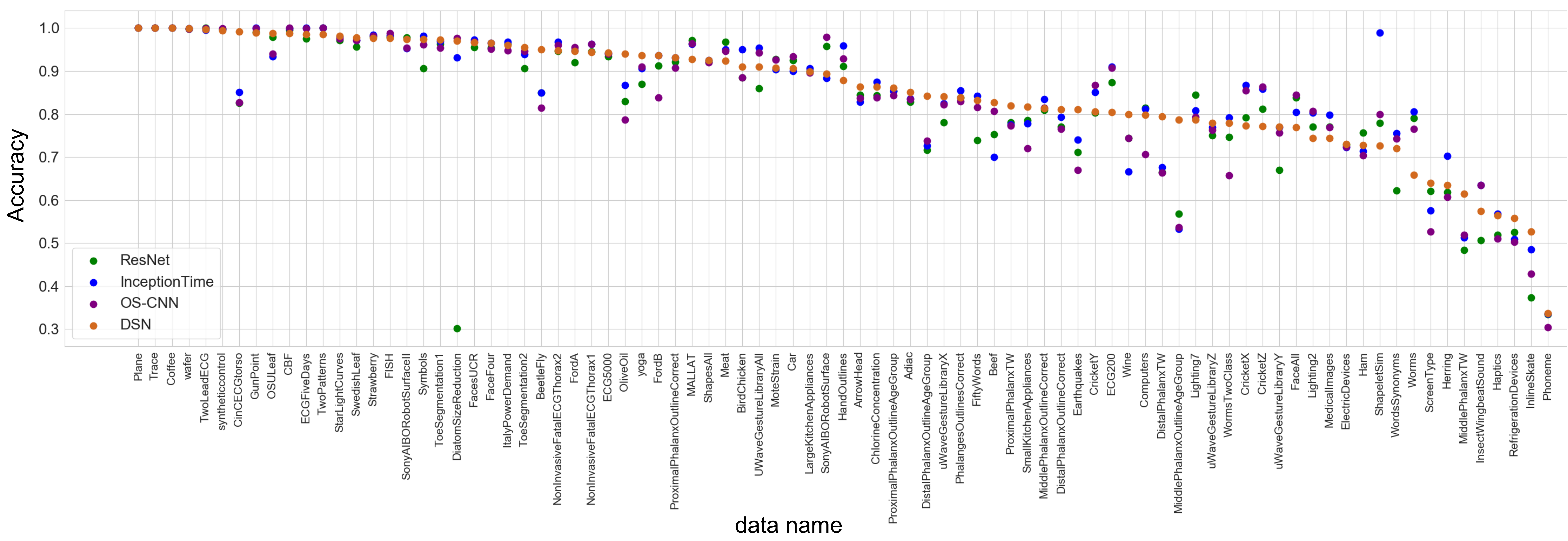}
    \caption{Visualizing classification accuracies on UCR 85 archive for the comparison of DSN and the baselines.}
    \label{fig:detailed_UCR}
\end{figure}

The detailed results for UCR 85 archive are given in Table~ \ref{tab:UCRresults_detailed}. We also visualize the detailed results in Figure~\ref{fig:detailed_UCR}, where we can see that the accuracies of DSN are nearly on the top in most of the datasets. In addition, it is worth noting that DSN loses by a slight margin in some datasets, but wins by a big margin in others. 

\section{More Implementation Details}
\label{appendix:D}

In our dynamic sparse training algorithm \ref{alg:DSN}, we choose the cosine annealing defined in Eq. \ref{eq:cosine annealing} with $\Delta T=5$ epochs and $\alpha=0.5$. For UEA 30 archive, in each dynamic sparse CNN layer, the output channel is set to 177 with sparsity ratio $S=90\%$.
For UCR 112 archive, except for each CNN layer is padded with zero-value, other hyper-parameters are the same as in UCR 85 archive.
We use PyTorch\footnote{https://pytorch.org/} to implement our method and run our experiments on Nvidia Tesla P100.

\begin{table}[!htb]
\caption{Test accuracies (ACC(\%)) for UCR 85 archive and resource cost (i.e. Params (K) and FLOPs (M)) of our method. Test accuracies of our method are run five times and reported with (mean$\pm$std).}
\centering
\resizebox{\linewidth}{!}{
\begin{tabular}{lccccrr}
\toprule[1pt]
 & ResNet & InceptionTime &  OS-CNN &  DSN (ours) & Params & FLOPs \\
\toprule[0.5pt] \midrule[0.5pt] 
50Words & 73.96 & 84.18 & 81.60 & 83.16$\pm$0.69 & 132.27 & 68.36 \\
Adiac & 82.89 & 83.63 & 83.45 & 85.06$\pm$0.73 & 130.41 & 44.68 \\
ArrowHead & 84.46 & 82.86 & 83.77 & 86.40$\pm$0.23 & 125.55 & 62.31 \\
Beef & 75.33 & 70.00 & 80.67 & 82.67$\pm$2.49 & 125.83 & 116.65 \\
BeetleFly & 85.00 & 85.00 & 81.50 & 95.00$\pm$0.00 & 125.40 & 126.97 \\
BirdChicken & 88.50 & 95.00 & 88.50 & 91.00$\pm$2.00 & 125.40 & 126.97 \\
CBF & 99.50 & 99.89 & 99.99 & 98.78$\pm$0.31 & 125.55 & 31.82 \\
Car & 92.50 & 90.00 & 93.33 & 90.67$\pm$2.26 & 125.69 & 143.14 \\
ChlorineConcentration & 84.36 & 87.53 & 83.87 & 86.33$\pm$0.20 & 125.55 & 41.24 \\
CinCECGtorso & 82.61 & 85.14 & 82.75 & 99.20$\pm$0.18 & 125.69 & 406.37 \\
Coffee & 100.00 & 100.00 & 100.00 & 100.00$\pm$0.00 & 125.40 & 70.95 \\
Computers & 81.48 & 81.20 & 70.68 & 79.84$\pm$0.54 & 125.40 & 178.53 \\
CricketX & 79.13 & 86.67 & 85.51 & 77.28$\pm$1.57 & 126.83 & 74.71 \\
CricketY & 80.33 & 85.13 & 86.72 & 80.51$\pm$1.04 & 126.83 & 74.71 \\
CricketZ & 81.15 & 85.90 & 86.31 & 77.13$\pm$1.85 & 126.83 & 74.71 \\
DiatomSizeReduction & 30.13 & 93.14 & 97.71 & 96.99$\pm$0.48 & 125.69 & 85.63 \\
DistalPhalanxOutlineAgeGroup & 71.65 & 72.66 & 73.81 & 84.20$\pm$0.33 & 125.55 & 19.92 \\
DistalPhalanxOutlineCorrect & 77.10 & 79.35 & 76.59 & 81.13$\pm$1.09 & 125.40 & 19.89 \\
DistalPhalanxTW & 66.47 & 67.63 & 66.40 & 79.40$\pm$0.37 & 125.98 & 20.00 \\
ECG200 & 87.40 & 91.00 & 90.80 & 80.40$\pm$0.49 & 125.40 & 23.85 \\
ECG5000 & 93.42 & 94.09 & 94.01 & 94.32$\pm$0.12 & 125.83 & 34.85 \\
ECGFiveDays & 97.48 & 100.00 & 99.95 & 98.51$\pm$0.48 & 125.40 & 33.77 \\
Earthquakes & 71.15 & 74.10 & 66.98 & 81.06$\pm$0.00 & 125.40 & 126.97 \\
ElectricDevices & 72.91 & 72.27 & 72.36 & 73.02$\pm$0.74 & 126.12 & 24.00 \\
FISH & 97.94 & 98.29 & 98.74 & 97.71$\pm$0.72 & 126.12 & 114.97 \\
FaceAll & 83.88 & 80.41 & 84.47 & 76.91$\pm$0.16 & 127.12 & 32.87 \\
FaceFour & 95.45 & 96.59 & 95.11 & 96.59$\pm$0.00 & 125.69 & 86.88 \\
FacesUCR & 95.47 & 97.32 & 96.74 & 96.62$\pm$0.23 & 127.12 & 32.87 \\
FordA & 92.05 & 94.83 & 95.48 & 94.64$\pm$0.16 & 125.40 & 124.00 \\
FordB & 91.31 & 93.65 & 83.79 & 93.62$\pm$0.67 & 125.40 & 124.00 \\
GunPoint & 99.07 & 100.00 & 99.93 & 98.93$\pm$0.53 & 125.40 & 37.24 \\
Ham & 75.71 & 71.43 & 70.38 & 72.76$\pm$1.66 & 125.40 & 106.89 \\
HandOutlines & 91.11 & 95.95 & 92.95 & 87.90$\pm$0.68 & 125.40 & 671.54 \\
Haptics & 51.88 & 56.82 & 51.01 & 56.43$\pm$2.30 & 125.83 & 270.82 \\
Herring & 61.88 & 70.31 & 60.78 & 63.44$\pm$4.49 & 125.40 & 126.97 \\
InlineSkate & 37.31 & 48.55 & 42.93 & 52.73$\pm$4.52 & 126.12 & 466.69 \\
InsectWingbeatSound & 50.65 & 63.48 & 63.52 & 57.49$\pm$0.52 & 126.69 & 63.77 \\
ItalyPowerDemand & 96.30 & 96.79 & 94.72 & 96.05$\pm$0.13 & 125.40 & 6.01 \\
LargeKitchenAppliances & 89.97 & 90.67 & 89.57 & 89.92$\pm$0.26 & 125.55 & 178.56 \\
Lighting2 & 77.05 & 80.33 & 80.66 & 74.43$\pm$2.45 & 125.40 & 157.95 \\
Lighting7 & 84.52 & 80.82 & 79.32 & 78.63$\pm$1.40 & 126.12 & 79.27 \\
MALLAT & 97.16 & 96.29 & 96.38 & 92.71$\pm$1.83 & 126.26 & 254.05 \\

\bottomrule[1pt]
 \end{tabular}
 }
\label{tab:UCRresults_detailed}
\end{table}

\begin{table}[!htb]
\centering
\resizebox{\linewidth}{!}{
\begin{tabular}{lccccrr}
\toprule[1pt]
 & ResNet & InceptionTime &  OS-CNN &  DSN (ours) & Params & FLOPs \\
\toprule[0.5pt] \midrule[0.5pt] 
Meat & 96.83 & 95.00 & 94.67 & 92.33$\pm$2.91 & 125.55 & 111.13 \\
MedicalImages & 77.03 & 79.87 & 76.95 & 74.39$\pm$0.24 & 126.55 & 24.82 \\
MiddlePhalanxOutlineAgeGroup & 56.88 & 53.25 & 53.64 & 78.65$\pm$0.77 & 125.55 & 19.92 \\
MiddlePhalanxOutlineCorrect & 80.89 & 83.51 & 81.41 & 81.33$\pm$1.27 & 125.40 & 19.89 \\
MiddlePhalanxTW & 48.44 & 51.30 & 51.95 & 61.45$\pm$0.40 & 125.98 & 20.00 \\
MoteStrain & 92.76 & 90.34 & 92.64 & 90.73$\pm$0.43 & 125.40 & 20.88 \\
NonInvasiveFatalECGThorax1 & 94.54 & 96.23 & 96.27 & 94.36$\pm$0.50 & 131.12 & 187.11 \\
NonInvasiveFatalECGThorax2 & 94.61 & 96.74 & 96.01 & 94.72$\pm$0.33 & 131.12 & 187.11 \\
OSULeaf & 97.85 & 93.39 & 94.01 & 98.84$\pm$0.31 & 125.98 & 106.01 \\
OliveOil & 83.00 & 86.67 & 78.67 & 94.00$\pm$3.27 & 125.69 & 141.40 \\
PhalangesOutlinesCorrect & 83.90 & 85.43 & 82.97 & 83.87$\pm$0.78 & 125.40 & 19.89 \\
Phoneme & 33.43 & 33.54 & 30.45 & 33.70$\pm$0.79 & 130.70 & 254.93 \\
Plane & 100.00 & 100.00 & 100.00 & 100.00$\pm$0.00 & 126.12 & 35.89 \\
ProximalPhalanxOutlineAgeGroup & 85.32 & 85.37 & 84.39 & 86.05$\pm$0.50 & 125.55 & 19.92 \\
ProximalPhalanxOutlineCorrect & 92.13 & 93.13 & 90.79 & 93.20$\pm$0.14 & 125.40 & 19.89 \\
ProximalPhalanxTW & 78.05 & 77.56 & 77.32 & 81.95$\pm$0.29 & 125.98 & 20.00 \\
RefrigerationDevices & 52.53 & 50.93 & 50.29 & 55.84$\pm$1.27 & 125.55 & 178.56 \\
ScreenType & 62.16 & 57.60 & 52.64 & 63.95$\pm$1.11 & 125.55 & 178.56 \\
ShapeletSim & 77.94 & 98.89 & 79.94 & 72.67$\pm$4.87 & 125.40 & 124.00 \\
ShapesAll & 92.13 & 92.50 & 92.03 & 92.50$\pm$0.61 & 133.70 & 128.63 \\
SmallKitchenAppliances & 78.61 & 77.87 & 72.08 & 81.76$\pm$0.40 & 125.55 & 178.56 \\
SonyAIBORobotSurface & 95.81 & 88.35 & 97.95 & 89.32$\pm$1.82 & 125.40 & 17.41 \\
SonyAIBORobotSurfaceII & 97.78 & 95.28 & 95.38 & 97.46$\pm$0.75 & 125.40 & 16.17 \\
StarLightCurves & 97.18 & 97.92 & 97.50 & 98.20$\pm$0.09 & 125.55 & 253.91 \\
Strawberry & 98.05 & 98.38 & 98.19 & 97.72$\pm$0.29 & 125.40 & 58.31 \\
SwedishLeaf & 95.63 & 97.12 & 97.12 & 97.79$\pm$0.19 & 127.26 & 32.16 \\
Symbols & 90.64 & 98.19 & 96.12 & 97.43$\pm$0.40 & 125.98 & 98.82 \\
ToeSegmentation1 & 96.27 & 96.93 & 95.39 & 97.28$\pm$0.33 & 125.40 & 68.72 \\
ToeSegmentation2 & 90.62 & 93.85 & 94.62 & 95.54$\pm$0.58 & 125.40 & 85.08 \\
Trace & 100.00 & 100.00 & 100.00 & 100.00$\pm$0.00 & 125.69 & 68.28 \\
TwoLeadECG & 100.00 & 99.56 & 99.92 & 99.70$\pm$0.07 & 125.40 & 20.38 \\
TwoPatterns & 99.99 & 100.00 & 100.00 & 98.50$\pm$0.26 & 125.69 & 31.84 \\
UWaveGestureLibraryAll & 85.95 & 95.45 & 94.25 & 90.99$\pm$0.38 & 126.26 & 234.46 \\
Wine & 74.44 & 66.67 & 74.44 & 80.00$\pm$2.96 & 125.40 & 58.06 \\
WordsSynonyms & 62.24 & 75.55 & 74.23 & 72.01$\pm$1.38 & 128.69 & 67.64 \\
Worms & 79.09 & 80.52 & 76.49 & 65.86$\pm$1.83 & 125.83 & 223.23 \\
WormsTwoClass & 74.68 & 79.22 & 65.71 & 77.90$\pm$1.35 & 125.40 & 223.15 \\
syntheticcontrol & 99.83 & 99.67 & 99.93 & 99.40$\pm$0.13 & 125.98 & 15.05 \\
uWaveGestureLibraryX & 78.05 & 82.47 & 82.18 & 84.06$\pm$0.39 & 126.26 & 78.31 \\
uWaveGestureLibraryY & 67.01 & 76.88 & 75.72 & 77.10$\pm$0.64 & 126.26 & 78.31 \\
uWaveGestureLibraryZ & 75.01 & 76.97 & 76.36 & 77.98$\pm$0.41 & 126.26 & 78.31 \\
wafer & 99.86 & 99.87 & 99.84 & 99.91$\pm$0.02 & 125.40 & 37.74 \\
yoga & 87.02 & 90.57 & 91.06 & 93.63$\pm$0.48 & 125.40 & 105.65 \\
\bottomrule[1pt]
 \end{tabular}
 }
\label{tab:UCRresults_detailed_2}
\end{table}

\begin{table}[!htb]
\caption{Test accuracies (ACC(\%)) for UEA 30 archive and resource cost (i.e. Params (K) and FLOPs (M)) of our method. Test accuracies of our method are run five times and reported with (mean$\pm$std).}
\centering
\resizebox{\linewidth}{!}{
\begin{tabular}{lccccrr}
\toprule[1pt]
 & MLSTM-FCN & TapNet &  OS-CNN &  DSN (ours) & Params & FLOPs \\
\toprule[0.5pt] \midrule[0.5pt] 
ArticularyWordRecognition & 97.30 & 98.70 & 98.75 & 98.40$\pm$0.25 & 146.35 & 41.58 \\
AtrialFibrillation & 26.70 & 33.30 & 23.33 & 6.67$\pm$0.00 & 142.00 & 180.42 \\
BasicMotions & 95.00 & 100.00 & 100.00 & 100.00$\pm$0.00 & 142.42 & 28.37 \\
CharacterTrajectories & 98.50 & 99.70 & 99.76 & 99.39$\pm$0.07 & 145.10 & 52.02 \\
Cricket & 91.70 & 95.80 & 99.31 & 98.89$\pm$0.56 & 143.85 & 338.23 \\
DuckDuckGeese & 67.50 & 57.50 & 54.00 & 56.80$\pm$3.25 & 221.60 & 119.04 \\
ERing & 13.30 & 13.30 & 88.15 & 92.22$\pm$1.12 & 142.66 & 18.55 \\
EigenWorms & 50.40 & 48.90 & 41.41 & 39.08$\pm$11.21 & 142.59 & 5075.34 \\
Epilepsy & 76.10 & 97.10 & 98.01 & 99.86$\pm$0.29 & 142.24 & 58.21 \\
EthanolConcentration & 37.30 & 32.30 & 24.05 & 24.49$\pm$0.89 & 142.24 & 493.68 \\
FaceDetection & 54.50 & 55.60 & 57.50 & 63.49$\pm$0.70 & 150.20 & 18.58 \\
FingerMovements & 58.00 & 53.00 & 56.75 & 49.20$\pm$1.17 & 143.36 & 14.32 \\
HandMovementDirection & 36.50 & 37.80 & 44.26 & 37.30$\pm$2.20 & 142.65 & 113.22 \\
Handwriting & 28.60 & 35.70 & 66.82 & 33.65$\pm$0.84 & 146.18 & 43.78 \\
Heartbeat & 66.30 & 75.10 & 48.90 & 78.34$\pm$0.79 & 145.30 & 117.00 \\
InsectWingbeat & 16.70 & 20.80 & 66.66 & 38.62$\pm$10.65 & 154.94 & 7.07 \\
JapaneseVowels & 97.60 & 96.50 & 99.12 & 98.86$\pm$0.11 & 143.66 & 8.53 \\
LSST & 37.30 & 56.80 & 41.25 & 60.26$\pm$4.53 & 144.21 & 10.66 \\
Libras & 85.60 & 85.00 & 95.00 & 96.44$\pm$0.27 & 144.15 & 13.22 \\
MotorImagery & 51.00 & 59.00 & 53.50 & 57.40$\pm$2.58 & 145.48 & 867.23 \\
NATOPS & 88.90 & 93.90 & 96.81 & 97.78$\pm$0.93 & 143.84 & 14.72 \\
PEMS-SF & 69.90 & 75.10 & 76.01 & 80.12$\pm$1.25 & 199.42 & 57.15 \\
PenDigits & 97.80 & 98.00 & 98.55 & 98.73$\pm$0.10 & 143.25 & 2.62 \\
PhonemeSpectra & 11.00 & 17.50 & 29.93 & 31.97$\pm$0.36 & 148.98 & 62.77 \\
RacketSports & 80.30 & 86.80 & 87.66 & 86.18$\pm$1.32 & 142.42 & 8.61 \\
SelfRegulationSCP1 & 87.40 & 65.20 & 83.53 & 71.74$\pm$0.79 & 142.06 & 252.93 \\
SelfRegulationSCP2 & 47.20 & 55.00 & 53.19 & 46.44$\pm$4.54 & 142.12 & 325.32 \\
SpokenArabicDigits & 99.00 & 98.30 & 99.66 & 99.10$\pm$0.21 & 143.90 & 23.85 \\
StandWalkJump & 6.70 & 40.00 & 38.33 & 38.67$\pm$4.99 & 142.12 & 705.04 \\
UWaveGestureLibrary & 89.10 & 89.40 & 92.66 & 91.56$\pm$0.44 & 142.95 & 89.08 \\
\bottomrule[1pt]
 \end{tabular}
 }
\label{tab:UCRresults_detailed_2}
\end{table}



\end{document}